\documentclass[11pt,letterpaper]{article}
\usepackage[margin=1in]{geometry}
\usepackage[T1]{fontenc}
\usepackage{lmodern}
\usepackage{amsmath,amssymb,amsthm,mathtools}
\usepackage{enumitem,booktabs,array,graphicx,longtable}
\usepackage[round]{natbib}
\usepackage[hidelinks]{hyperref}
\usepackage[section]{placeins}
\usepackage{microtype}
\hypersetup{pdftitle={Conformal Selection for Hierarchical Streaming Data},pdfauthor={Zhenyuan Sun, Chengyao Yu, Ruixing Ming, Bingyi Jing},pdfsubject={OHS and feedback-enhanced hierarchical selection: theory and experiments}}
\newtheorem{assumption}{Assumption}
\newtheorem{theorem}{Theorem}[section]

\newtheorem{proposition}[theorem]{Proposition}

\theoremstyle{definition}

\newtheorem{remark}[theorem]{Remark}
\newcommand{\ind}{\mathbf 1}
\newcommand{\E}{\mathbb E}
\newcommand{\Prob}{\mathbb P}
\newcommand{\calZ}{\mathcal Z}
\newcommand{\calN}{\mathcal N}
\newcommand{\calK}{\mathcal K}
\newcommand{\calR}{\mathcal R}

\newcommand{\FDR}{\operatorname{FDR}}

\newcommand{\Ed}{\overset{\mathrm d}{=}}

\title{Conformal Selection for Hierarchical Streaming Data}
\author{Zhenyuan Sun$^\text{a}$\thanks{Co-first authors.}, Chengyao Yu$^\text{b}$\footnotemark[1], Ruixing Ming$^\text{c}$\thanks{Corresponding author.},  Bingyi Jing$^\text{d,e}$\\
$^\text{a}$School Mathematical Sciences, Peking University;\\
$^\text{b}$Department of Statistics and Data Science,\\
Southern University of Science and Technology;\\
$^\text{c}$School of Statistics and Data Science,
Zhejiang Gongshang University;\\
$^\text{d}$School of Artificial Intelligence,
The Chinese University of Hong Kong, Shenzhen;\\
$^\text{e}$Shenzhen Loop Area Institute
}
\date{}
\newcounter{algorithm}
\newenvironment{algorithmblock}[1]{  \par\medskip\noindent\begin{minipage}{\textwidth}  \refstepcounter{algorithm}  \hrule width\textwidth\vskip4pt  \noindent Algorithm~\thealgorithm: #1\par
  \vskip4pt\hrule width\textwidth\vskip5pt  \normalsize
}{  \par\vskip5pt\hrule width\textwidth
  \end{minipage}\par\medskip
}

\usepackage{float}
\begin{document}
\maketitle
\begin{abstract}
Many real-world tasks involve making sequential decisions about whether to select candidates before their responses of interest are revealed. In this work, we address the online selection problem in the context of  hierarchical data, where candidates arrive sequentially from different populations. This extends existing online conformal selection frameworks beyond the standard i.i.d./exchangeable data assumption.
The primary challenge lies in constructing valid statistics for each candidate sequentially under the hierarchical data structure.
To this end, we first develop \textit{Online Hierarchical Selection} (OHS), a distribution-free method
that enables principled candidate selection for hierarchical data streams. Specifically, we introduce methods to construct rank-based e-values for both between-group and within-group samples, and apply averaged e-values to the online e-BH procedure. We then develop OHS with feedback (OHSF), which boosts the power of OHS by leveraging potential response feedback.
Theoretically, the proposed methods achieve false discovery rate control under mild hierarchical exchangeability and data arrival assumptions at any time.
We also generalize methods to broader settings, such as data with covariate shift and selection under effect-related responses.
Extensive experiments on both simulated and real-world datasets demonstrate the effectiveness of the proposed methods.
\end{abstract}

\section{Introduction}\label{sec:intro}
{
Many scientific discovery and decision-making tasks involve selecting candidates with desired outcomes sequentially from a data stream before their outcomes are observed. This paper focuses such online selection problem in the context of hierarchical data, where  the streaming candidates can be organized into groups and i.i.d./exchangeability only holds within and across groups. Such data structure is common in reality:
\begin{itemize}
\item In \textit{multi-center clinical studies}, patients are grouped by hospitals or study sites, which may have distinct populations and outcome distributions~\citep{higgins2009}.
\item In \textit{cluster-randomized trials}, individuals are nested within villages, cities, or schools and share a cluster-specific treatment or environment~\citep{su2021model}.
\item  In \textit{repeated assessment}, the same data point may receive multiple ratings or annotations, creating observations nested within the assessed unit~\citep{stutz2023conformal}.
\end{itemize}
While selections concern spending scarce resources such as experimental capacity, follow-up budgets, and patient slots, we aim to develop principled online selection procedures that rigorously control false selections under the hierarchical steaming data.

The fast-developing artificial intelligence presents great promise in assisting decision-making due to its excellent predictive power across a wide range of fields~\citep{dara2022machine,watson2023novo,price2025probabilistic}. For the online selection task, a promising way is to leverage machine-predicted outcomes to select those desired candidates. However, such decisions are commonly based on black-box predictive models, which can not control false selections.
To achieve principled selection, conformal selection (CS)~\citep{jin2023} provides a model-agnostic framework for selecting candidates with large outcomes in offline settings, which converts predictive scores into conformal p-values and applies the Benjamini-Hochberg (BH) procedure \citep{benjamini1995} to select candidates with false discovery rate (FDR) control. Numerous studies have extended CS to different settings~\citep{gui2024conformal,gui2025,bai2025,yu2026robust,jin2026model}. Recently, to tackle the online setting where candidates arrive sequentially, \citep{liu2026} developed an online extension of CS by applying the online BH procedure~\citep{xu2024} to conformal p-values. Other works, including \citep{fischer2026,lu2026}, further generalized the online selection method and extended to settings with feedback. 

Despite these remarkable advancements in principled candidate selection, none of these works address the selection problem under hierarchical streaming data.
The statistical challenge is that these methods mainly rely on the i.i.d./exchangeable data assumption, and the failure of this assumption will directly affect the validity of test statistics, such as conformal p-values. One of the most related works is \citep{lee2025}, which generalizes CS~\citep{jin2023} to hierarchical data. Specifically, they introduce methods to construct subsampling/hierarchical conformal e-values and subsequently apply the e-BH procedure~\citep{wang2022} to provide FDR control, where the e-value is a competitor of the p-value and has been widely applied to hypothesis testing with complex data structures~\citep{yu2024generalized, yu2026anytime, gablenz2025catch}. However, their method is designed for offline settings, and their statistics can not be directly applied to online selection procedures, such as online e-BH.
Extending conformal selection for hierarchical streaming data therefore needs novel techniques.
This paper aims to address this gap.


In this paper, we develop online selection frameworks for hierarchical streaming data, covering both settings with and without response feedback.
For the no-feedback setting, we propose \textit{Online Hierarchical Selection} (OHS), which constructs two complementary rank-based e-values for each arriving candidate. Specifically, we construct a within-group e-value obtained by comparing the threshold score of the candidate with null calibration scores from the same group, and a between-group e-value by comparing it with one uniformly sampled null representative from each   group other than the candidate group.
The two e-values are combined through a prespecified weighted average, allowing OHS to leverage information from both within- and between-group comparisons.
We then apply online e-BH to the combined e-values and progressively expand the selection set as new candidates arrive, without revoking previous selections.
For settings requiring immediate and irreversible decisions, we further develop Online Hierarchical Selection with Feedback (OHSF). After each decision, we incorporate newly revealed null observation into the calibration pool of the corresponding group, regardless of the selection outcome, and apply randomized e-LOND~\citep{xu2024} to the resulting hierarchical e-values for subsequent decisions.
We also extend the proposed frameworks to accommodate covariate shift, effect-related responses, and the arrival of previously unseen groups.

Theoretically, we establish finite-sample FDR control for both OHS and OHSF under the corresponding assumptions. For OHS, we establish the validity of the within-group and between-group rank-based e-values under score monotonicity.
For OHSF, we establish FDR control for immediate decisions based on feedback-updated calibration pools, without requiring the score function to be monotone in the response. We further establish FDR guarantees for extensions to covariate shift, effect-related responses, and previously unseen groups under their respective conditions.

}

{

}

{
The rest of the paper is organized as follows.
Section~\ref{sec:setup} formulates the problem and introduces the data assumptions.
Section~\ref{sec:ohs} develops the OHS and OHSF, and establishes their theoretical results.
Section~\ref{sec:extensions} extends the proposed methods to settings, including data with covariate shift, selection under effect-related responses, and selection for unseen groups.
Simulation results and real data applications are presented in Sections~\ref{sec:simulation} and \ref{sec:case}, respectively.
Section~\ref{sec:conclusion} provides the discussions.
All proofs are provided in appendix.
}

\section{Problem Setup}\label{sec:setup}
{
Consider the online setting where data from $K\geq 1$ groups arrives sequentially.
For concise, we consider finite online candidates and let $T$ be an upper bound.
For the $t$-th candidate, let
$X_t\in\mathcal X\subseteq\mathbb R^d$ denote its feature vector, $Y_t\in\mathbb R$ denote its response, and $L_t\in[K]$ represent its group label.
Consider that we have $T_0$ calibration data $\{(X_t,Y_t,L_t)\}_{-T_0+1\leq t\leq 0}$. The test sample $(X_t,Y_t,L_t)$ arrives sequentially for $t=1,\dots,T$, where the response $Y_t$ is unobservable at time $t$.

For the considered online setting, we study the task of selecting candidates with large responses~\citep{jin2023}.
{Specifically, let $c:\mathcal X\to\mathbb R$ be a prespecified measurable threshold function. We aim to select candidates satisfying $Y_t\ge c(X_t)$ sequentially for $1\leq t\leq T$. A constant threshold is included as a special case.}
{Write $\theta_i=\ind\{Y_i\ge c(X_i)\}$ for both calibration and online observations}, and let $\calR_t\subseteq\{1,\ldots,t\}$ be the set selected at time $t$, $1\leq t\leq T$.
For a nominal level $\alpha\in(0,1)$, our goal is to determine $\{\mathcal{R}_t\}_{1\leq t\leq T}$ such that
\[
 \FDR(t)=\E\!\left[
 \frac{\sum_{i\in\calR_t}(1-\theta_i)}{|\calR_t|\vee1}
 \right]\le\alpha,\quad 1\leq t\leq T.
\]
We first consider the non-feedback setting, i.e., $\{Y_t\}_{1\leq t \leq T}$ is always unobservable.
For this setting, we focus on the practical \textit{Accept-to-Reject Changes} (ARC): non-selected candidates can be reconsidered for selection later, and the decision of selection is irreversible. That is, the selection sets are nested: $\varnothing=\calR_0\subseteq\calR_1\subseteq\cdots\subseteq\calR_T$.
We also study the selection problem under immediate feedback, where $\theta_t$ is revealed at time $t+1$.
For this case, we consider a quite different setup, where each candidate is considered for selection only once at arrival.
}

{
\subsection{Hierarchical Streaming Data}
We now formulate the structure of hierarchical streaming data.
For clarity, we first re-index the observations within each group.
Specifically, let $\calZ=(Z_{k,j})_{k\in[K],\,1\le j\le n_k}$, where $Z_{k,j}=(X_{k,j},Y_{k,j})$ represents the $j$-th observation from group $k$ and $n_k$ is the total number of possible observations in that group across $\{(X_t,Y_t,L_t)\}_{-T_0+1\leq t\leq T}$.
Let $j(t)=\sum_{u=-T_0+1}^{t}\ind\{L_u=L_t\}$ counts the observations from group $L_t$ up to time $t$. Then we have $(X_t,Y_t)=Z_{L_t,j(t)}$.
The following assumption describes the hierarchical exchangeability,
}

{
\begin{assumption}[Hierarchical exchangeability]\label{ass:he}
The joint law of $\calZ$ satisfies:
\begin{enumerate}[label=\arabic*.,leftmargin=*]
\item \textbf{Within-group exchangeability.} For every $k$ and every permutation $\pi$ of $\{1,\ldots,n_k\}$ that leaves all other groups fixed,
\begin{equation}
\begin{aligned}
 &\quad~(Z_{1,1},\ldots,Z_{k-1,n_{k-1}};Z_{k,1},\ldots,Z_{k,n_k};Z_{k+1,1},\ldots,Z_{K,n_K})\\
 &\Ed(Z_{1,1},\ldots,Z_{k-1,n_{k-1}};Z_{k,\pi(1)},\ldots,Z_{k,\pi(n_k)};Z_{k+1,1},\ldots,Z_{K,n_K}).
\end{aligned}
\end{equation}
\item \textbf{Between-group exchangeability.} For every permutation $\sigma$ of $\{1,\ldots,K\}$,
\begin{equation}
 \begin{aligned}
 &\quad~(Z_{1,1},\ldots,Z_{1,n_1};\ldots;Z_{K,1},\ldots,Z_{K,n_K})\\
 &\Ed
 (Z_{\sigma(1),1},\ldots,Z_{\sigma(1),n_{\sigma(1)}};
 \ldots;Z_{\sigma(K),1},\ldots,Z_{\sigma(K),n_{\sigma(K)}}).
 \end{aligned}
\end{equation}
\end{enumerate}
\end{assumption}
The group sizes $n_k$ need not be independent of the observations
within their groups. In between-group block exchangeability, $n_k$ move together with complete blocks under permutation. Assumption~\ref{ass:he} is weaker than global individual exchangeability.
We retain the hierarchical exchangeability requirement adopted in theLee and Ren~\citep[Assumption 1, item 1]{lee2025}. 
For the arrival order, we impose the following assumption.

\begin{assumption}[Content-free arrival order]\label{ass:flow}
For the fixed calibration size $T_0$ and horizon $T$,
the group sequence $(L_i)_{-T_0+1\le i\le T}$ are independent of the entire array $\calZ$:
\[
 (L_i)_{-T_0+1\le i\le T}\ \perp\!\!\!\perp\ \calZ.
\]
\end{assumption}

 Assumption~\ref{ass:flow} requires the group sequence to be independent of the underlying data, so that neither the calibration allocation nor the online arrival order depends on features, responses, or group sizes. Together with hierarchical exchangeability, this preserves the symmetry needed for valid rank-based evidence construction.
 }

\section{Methodology}\label{sec:ohs}
{
In Section~\ref{sec:OHSm}, we first develop OHS, by introducing methods to construct within-group and between-group e-values. Section~\ref{sec:gaif} then develops OHSF, which considers the immediate decisions and extends OHS to the feedback setting.

\subsection{Online Hierarchical Selection}\label{sec:OHSm}
Let $V:\mathcal X\times\mathbb R\to\mathbb R$ be a given
nonconformity score function such that $V(x,y)\leq V(x,y')$ holds for any $x\in\mathcal{X}$ and any $y,y'\in\mathbb{R}$ satisfying $y\leq y'$.
Therefore, a smaller ${V(X_t,c(X_t))}$ provides stronger evidence against ${Y_t<c(X_t)}$.
For example, we may take $V(x,y)=y-\widehat\mu(x)$, where
$\widehat\mu$ is a given predictor of $\E[Y\mid X=x]$.
Let $V_t=V (X_t,Y_t)$ and $\hat{V}_t={V(X_t,c(X_t))}$, where $V_t$ is not observable for $1\leq t\leq T$.
Generally speaking, the key step of OHS is to construct valid within-group and between-group e-values by leveraging ranking information about nonconformity scores, where each e-value $E_t$ satisfies 
\begin{equation}\label{eq:joint-e}
 E_t\ge0,\qquad \E\!\left[E_t(1-\theta_t)\right]\le1.
\end{equation}
The OHS then combines them through a weighted average and applies online e-BH to control FDR. 
In what follows, we describe constructions of both within-group e-values and between-group e-values.

}

\paragraph{Within-group e-values.}
{Let $\calN_k$ denote the set of null calibration indices in group $k$:
\[
 {\calN_k=\{i\le0:L_i=k, Y_i<c(X_i)\}},\quad k=1,\dots,K.
\]
Write $m_k=|\calN_k|$.
Let $q_w\in(0,1]$ be a prespecified rank fraction that determines the cutoff.
We construct the within-group e-value by
\begin{equation}\label{eq:within}
 E_{t,w}=\frac{m_{L_t}+1}{\lceil q_w(m_{L_t}+1)\rceil}
 \ind\!\left\{1+\sum_{i\in\calN_{L_t}}\ind\{V_i<\widehat V_t\}
 \le\lceil q_w(m_{L_t}+1)\rceil\right\}.
\end{equation}
If $\calN_\ell$ is empty, we set $E_{t,w}=0$. 
}

{
\begin{theorem}[Within-group validity]\label{lem:within}
Under Assumptions~\ref{ass:he}--\ref{ass:flow} and the monotonic score $V$, the fixed-calibration within-group e-value satisfies $\E\left[E_{t,w}(1-\theta_t)\right]\le1$.
\end{theorem}
Theorem~\ref{lem:within} demonstrates the validity of the within-group e-values. However, the within-group e-value uses only information from the group of the observation arriving at time $t$, which is insufficient to exploit the full structure of hierarchical data. In what follows, we introduce the construction of between-group e-values, which exploits information from different groups.
}

{
\paragraph{Between-group e-values.}
Let $ \calK_t=\{k\ne L_t:m_k>0\}$ denote the set of groups other than $L_t$ that contain at least one null calibration observation.
For each $k\in\calK_t$, we independently draw an index $i_{k,t}$ uniformly from $\calN_k$.
Let $q_b\in(0,1]$ be a prespecified rank fraction. The between-group e-value is defined as
\begin{equation}\label{eq:between}
 E_{t,b}=\frac{1+|\calK_t|}{\min\{\lceil q_b K\rceil,1+|\calK_t|\}}
 \ind\!\left\{1+\sum_{k\in\calK_t}\ind\{V_{i_{k,t}}<\widehat V_t\}
 \le\lceil q_b K\rceil\right\}.
\end{equation}
If $\calK_t$ is empty, we set $E_{t,b}=0$. Theorem~\ref{thm:between} demonstrates the validity of between-group e-values.
\begin{theorem}[Between-group validity]\label{thm:between}
Let $E_{t,b}$ defined by \eqref{eq:between}.
Under Assumptions~\ref{ass:he}--\ref{ass:flow}, and score monotonicity, we have $\E\!\left[E_{t,b}(1-\theta_t)\right]\le1$.
\end{theorem}
}

{
\paragraph{Averaged e-values.}
We combine the within-group and between-group e-values through a prespecified weighted average:
\begin{equation}\label{eq:mixture}
E_t=a_tE_{t,w}+(1-a_t)E_{t,b},
\end{equation}
where the mixture weight is specified as
\begin{equation}\label{eq:weight}
a_t=\frac{T_0+t-1}{T_0+t-1+K^2}.
\end{equation}
The weight depends only on the calibration size $T_0$, the arrival time $t$, and the number of groups $K$, and is therefore determined independently of the online data. We compute $E_t$ upon arrival and retain it unchanged for subsequent applications of online e-BH. Further discussion of the choices of $a_t$, $q_b$, and $q_w$ is provided in Remark~\ref{remark_34}.
}
{
\paragraph*{Applying online e-BH procedure.}
At each time $t$, we apply the online e-BH procedure of \citep{fischer2026}.
Fix a target FDR level $\alpha\in(0,1)$ and a prespecified sequence of nonnegative weights $\gamma_i$ satisfying $\sum_i\gamma_i\le1$.
These weights allocate the testing budget across candidates, with larger weights corresponding to lower evidence thresholds. We compute $k_t^*$ by
\begin{equation}\label{eq:online-ebh}
 k_t^*=\max\left\{k\in[t]:\sum_{i\le t}\ind\!\left\{E_i\ge\frac1{\alpha\gamma_i k}\right\}\ge k\right\},
\end{equation}
and determine the selection set $\mathcal{R}_t$ as
\begin{equation}\label{eq:online-ebh-set}
 \calR_t=\left\{i\le t:E_i\ge\frac1{\alpha\gamma_i k_t^*}\right\}.
\end{equation}
If $k_t^*$ does not exist, we set $\calR_t=\emptyset$.
}

{
\begin{theorem}\label{thm:ohs}
Let $E_t$ and $\mathcal{R}_t$ be defined by \eqref{eq:mixture} and \eqref{eq:online-ebh-set}, respectively.
Under Assumptions~\ref{ass:he}--\ref{ass:flow} and score monotonicity, then $E_t$ satisfies~\eqref{eq:joint-e}. Then we have $\FDR(t)\le\alpha$, $t\geq 1$.
\end{theorem}
}

\begin{algorithmblock}{Online hierarchical selection (OHS-S)}\label{alg:ohs}
\noindent\textbf{Require:} Calibration null pools $\{\calN_k\}_{k=1}^K$,
score function $V$, {threshold function $c(\cdot)$}, calibration size $T_0$,
FDR level $\alpha$, rank fractions $q_w$ and $q_b\le 1/K$,
and deterministic spending sequence $(\gamma_t)_{t\geq1}$ with $\sum_t\gamma_t\le1$.
\begin{enumerate}[label=\arabic*:,leftmargin=2em,labelsep=.5em,nosep]
\item \textbf{for} $t=1,2,\ldots$ \textbf{do}
\item \hspace*{1em}\parbox[t]{\dimexpr\linewidth-1em\relax}{Observe $(X_t,L_t)$ and compute $\widehat V_t=V(X_t,c(X_t))$.
}
\item \hspace*{1em}\parbox[t]{\dimexpr\linewidth-1em\relax}{Compute the within-group e-value $E_{t,w}$ by~\eqref{eq:within}.}
\item \hspace*{1em}\parbox[t]{\dimexpr\linewidth-1em\relax}{For each $k\in\mathcal K_t$, independently draw
$i_{k,t}\sim\mathrm{Uniform}(\mathcal N_k)$
and compute $E_{t,b}$ using~\eqref{eq:between}.}
\item \hspace*{1em}\parbox[t]{\dimexpr\linewidth-1em\relax}{Compute $a_t$ by~\eqref{eq:weight} and store the combined e-value $E_t$ from~\eqref{eq:mixture}.}
\item \hspace*{1em}\parbox[t]{\dimexpr\linewidth-1em\relax}{Apply online e-BH using~\eqref{eq:online-ebh}--\eqref{eq:online-ebh-set} and report $\calR_t$.}
\item \textbf{end for}
\end{enumerate}
\noindent\textbf{Ensure:} Cumulative selection sets $\calR_1,\calR_2,\ldots$.
\end{algorithmblock}

{
\begin{remark}[Choice of parameters]\label{remark_34}
The FDR guarantee holds for any prespecified sequence $a_t\in[0,1]$. However, the choice of $a_t$ in \eqref{eq:weight} provides a convenient default. When the calibration sample is balanced and $t$ is small relative to $T_0$, it is approximately $(T_0/K)/(T_0/K+K)$, reflecting the relative sizes of the average calibration sample per group and the number of groups.
The rank fractions $q_w$ and $q_b$ control the trade-off between the magnitude and frequency of nonzero e-values: smaller values yield larger nonzero e-values but make them less frequent.
Theorem~\ref{thm:between} establishes validity for any prespecified $q_b\in(0,1]$, with $q_b=1/K$ corresponding to a rank-one cutoff.
In practice, we may tune $a_t$, $q_w$, and $q_b$ over a small grid using separate pilot data, and fix them before calibration and online selection.
\end{remark}

We next discuss three refinements of OHS.
The first two, presented in Remarks \ref{prop:average} and \ref{prop:center}, concern the construction of the between-group e-value, while the third modifies the selection rule.
}

{
\begin{remark}[Exact averaging]\label{prop:average}
Let $\mathcal G_t=
\sigma(
L_t,\widehat V_t,
\left(\mathcal N_k,(V_i)_{i\in\mathcal N_k}\right)_{k=1}^K
)$.
Conditional on $\mathcal G_t$, the representative indices
are independently and uniformly sampled from their respective
null calibration pools. Following Ren and Barber~\citep{renbarber2024},
we average over the representative sampling:
\begin{equation}\label{eq:rb}
\overline E_{t,b}
=
\frac{1+|\calK_t|}
{\min\{\lceil q_bK\rceil,1+|\calK_t|\}}
\Prob\!\left(
1+\sum_{k\in\calK_t}
\ind\{V_{i_{k,t}}<\widehat V_t\}
\le\lceil q_bK\rceil
\,\middle|\,\mathcal G_t
\right).
\end{equation}
We set $\overline E_{t,b}=0$ when $\calK_t$ is empty.
The conditional probability can be computed by convolving
the Bernoulli distributions of the group-wise comparisons.
This averaging removes the randomness due to representative
sampling and preserves \eqref{eq:joint-e}.
\end{remark}
}

{
\begin{remark}[Centering]\label{prop:center}
Fix a deterministic integer $u\ge0$ that does not exceed the calibration sample size in any group. For each group $k$, let $\mathcal P_k(u)=\{1,\ldots,u\}$ denote the indices of the first $u$ calibration observations, and let $\mathcal P_{k,0}(u)=\{j\in\mathcal P_k(u):\theta_{k,j}=0\}$. Define $\widehat\mu_k=|\mathcal P_{k,0}(u)|^{-1}\sum_{j\in\mathcal P_{k,0}(u)}V_{k,j}$ if $\mathcal P_{k,0}(u)\ne\emptyset$, using the mean of all scores in the prefix if $\mathcal P_{k,0}(u)=\emptyset$ and $u>0$, and setting $\widehat\mu_k=0$ when $u=0$. For subsequent observations $j>u$, define
\begin{equation}\label{eq:prefix}
V'_{k,j}=V_{k,j}-\widehat\mu_k,\qquad
\widehat V'_{k,j}=\widehat V_{k,j}-\widehat\mu_k.
\end{equation}
The first $u$ calibration observations are excluded only from between-group comparisons. The same validity guarantee holds for fixed, group-symmetric prefix adjustments that preserve monotonicity in the response \citep{vovk2005}.
\end{remark}

\paragraph{Uniform randomization.}
We further consider a randomized selection rule. Let
$U_t\overset{\mathrm{iid}}\sim\mathrm{Uniform}(0,1)$ be
independent of the data and all other randomization
\citep{xuramdas_random2023}, and define
$E_{t,U}=E_t/U_t$. We apply online e-BH to
$(E_{t,U})_{t\ge1}$ in place of $(E_t)_{t\ge1}$.
Although $E_{t,U}$ need not itself be an e-value, the
following result establishes FDR control and shows that
randomization can only enlarge the selection set.
Let $\mathcal R_t^{\mathrm{unboosted}}$ and
$\mathcal R_t^{\mathrm{boosted}}$ denote the selection sets
obtained using $(E_i)_{i\le t}$ and $(E_i/U_i)_{i\le t}$,
respectively.

\begin{proposition}[Randomized OHS]\label{thm:boost}
If $(E_t)_{t\ge1}$ satisfies \eqref{eq:joint-e}, then
online e-BH applied to $(E_t/U_t)_{t\ge1}$ satisfies
$\operatorname{FDR}_U(t)\le\alpha$ for each fixed $t$.
Moreover, we have $\mathcal R_t^{\mathrm{unboosted}}
\subseteq \mathcal R_t^{\mathrm{boosted}}$.
\end{proposition}
}

\subsection{Online Hierarchical Selection with Feedback}\label{sec:gaif}
{
We now focus on the state-feedback setting of Lu et al.~\citep{lu2026}.
At time $t$, we observe $(X_t,L_t)$ and must make an irreversible decision $\delta_t$. Then the ground truth of $\theta_t$ is revealed before $(X_{t+1},L_{t+1})$ arrives.
This feedback allows us to update the group-specific calibration pools using all previously revealed null observations, regardless of whether they were selected. Specifically, at time $t$, the null calibration pool for group $k$ is
\begin{equation}\label{eq:feedback-pools}
\calN_{k,t}=\calN_k\cup\{i<t:L_i=k,\theta_i=0\},\qquad
m_{k,t}=|\calN_{k,t}|.
\end{equation}

Specifically, we use the same prespecified feature score $\widehat V_i={V(X_i,c(X_i))}$
for candidates and calibration observations, so no monotonicity of
$V$ in the response is required. We break ties using independent
continuous random marks, generated once per observation and
reused across comparisons.
The within-group e-value is defined as
\begin{equation}\label{eq:p-within}
 E_{t,w}=
 \frac{m_{L_t,t}+1}{\lceil q_w(m_{L_t,t}+1)\rceil}
 \ind\!\left\{1+\sum_{i\in\calN_{L_t,t}}\ind\{\widehat V_i<\widehat V_t\}
 \le\lceil q_w(m_{L_t,t}+1)\rceil\right\}.
\end{equation}
We set $E_{t,w}=0$ when $\calN_{L_t,t}$ is empty.
For the between-group e-values, let $\calK_t^{\mathrm{fb}}
=\{k\ne L_t:m_{k,t}>0\}$ denote the groups other than $L_t$ with nonempty feedback calibration pools.
We construct $\overline E_{t,b}$ by applying the exact-averaging construction in \eqref{eq:rb}, with $\calN_k$ and $\calK_t$ replaced by $\calN_{k,t}$ and $\calK_t^{\mathrm{fb}}$, respectively, and using $\widehat V_i$ for both candidates and calibration observations. We set $\overline E_{t,b}=0$ when $\calK_t^{\mathrm{fb}}$ is empty.
When using the centering adjustment in \eqref{eq:prefix}, we exclude the prefix observations from between-group comparisons and use the adjusted scores for both candidates and calibration observations.
We then compute the averaged e-value by
\begin{equation}\label{eq:feedback-average}
 \overline E_t=a_tE_{t,w}+(1-a_t)\overline E_{t,b},
\end{equation}
where $a_t$ is defined in \eqref{eq:weight}.
Finally, fix a target FDR level $\alpha\in(0,1)$ and a prespecified spending sequence $\gamma_t\ge0$ satisfying $\sum_t\gamma_t\le1$. We draw $U_t\sim U(0,1)$ independently and apply randomized e-LOND directly to the combined e-values. The decision at time $t$ is given by
\begin{equation}\label{eq:gaif-r}
 \delta_t=\ind\{U_t\le\alpha_{t,R}\overline E_t\},\quad  \alpha_{t,R}=\min\{\alpha,\alpha\gamma_t[1+|\calR_{t-1}|]\}.
\end{equation}
Algorithm~\ref{alg:feedback} summarizes the procedure.
\begin{theorem}\label{thm:gaif-fdr}
Under Assumptions~\ref{ass:he}--\ref{ass:flow} and the feedback protocol above, rule~\eqref{eq:gaif-r} controls $\FDR(t)\le\alpha$ at every fixed $t$.
\end{theorem}
}

\begin{algorithmblock}{Online hierarchical selection with feedback (OHSF)}\label{alg:feedback}
\noindent\textbf{Require:} Initial calibration null pools $\{\calN_k\}_{k=1}^K$,
fixed score function $V$, {threshold function $c(\cdot)$}, calibration size $T_0$,
FDR level $\alpha$, rank fractions $q_w,q_b$, and deterministic spending
sequence $(\gamma_t)$ with $\sum_t\gamma_t\le1$.
\begin{enumerate}[label=\arabic*:,leftmargin=2em,labelsep=.5em,nosep]
\item Initialize $\calR_0=\varnothing$ and the pools $\calN_{k,1}$ using~\eqref{eq:feedback-pools}.
\item \textbf{for} $t=1,2,\ldots$ \textbf{do}
\item \hspace*{1em}\parbox[t]{\dimexpr\linewidth-1em\relax}{Observe $(X_t,L_t)$ and compute $\widehat V_t$.}
\item \hspace*{1em}\parbox[t]{\dimexpr\linewidth-1em\relax}{Compute $E_{t,w}$ using~\eqref{eq:p-within} and
$\overline E_{t,b}$ using~\eqref{eq:rb}.}
\item \hspace*{1em}\parbox[t]{\dimexpr\linewidth-1em\relax}{Compute $a_t$ using~\eqref{eq:weight} and combine the evidence using~\eqref{eq:feedback-average}.}
\item \hspace*{1em}\parbox[t]{\dimexpr\linewidth-1em\relax}{Draw $U_t$ independently from $\mathrm{Uniform}(0,1)$ and compute $\alpha_{t,R}$ and $\delta_t$ using~\eqref{eq:gaif-r}. Add $t$ to $\calR_{t-1}$ to form $\calR_t$ if $\delta_t=1$; otherwise retain $\calR_{t-1}$.}
\item \hspace*{1em}\parbox[t]{\dimexpr\linewidth-1em\relax}{After the decision, observe $\theta_t$ regardless of selection and update the pools for the next arrival using~\eqref{eq:feedback-pools}.}
\item \textbf{end for}
\end{enumerate}
\noindent\textbf{Ensure:} Decisions $\delta_1,\delta_2,\ldots$ and cumulative selection sets $\calR_1,\calR_2,\ldots$.
\end{algorithmblock}

\section{Extensions}\label{sec:extensions}

{
This section extends online selection of hierarchical data to settings including covariate shift, effect-related responses, and arrivals from previously unseen groups.}

\subsection{Selection under Covariate Shift between Calibration and Screening}
{
We consider the setting where the calibration dataset $\{(X_t,Y_t,L_t)\}_{-T_0+1\leq t\leq 0}$ has a covariate shift with the test points. Specifically, for each $k\in[K]$,}
\begin{equation}\label{eq:phase-shift-model}
\begin{aligned}
 (X_s)_{-T_0<s\le0,\,L_s=k}
 &\overset{\mathrm{i.i.d.}}{\sim}P_X^{\rm cal},\\
 (X_s)_{1\le s\le T,\,L_s=k}
 &\overset{\mathrm{i.i.d.}}{\sim}P_X^{\rm tar},\\
 Y_s\mid X_s
 \sim P_{Y\mid X=X_s},
 \quad \text{for }&-T_0<s\le T\text{ with }L_s=k.
\end{aligned}
\end{equation}
The arrival order continues to satisfy Assumption~\ref{ass:flow}. Assume $P_X^{\rm tar}\ll P_X^{\rm cal}$ and that the density ratio
\begin{equation}\label{eq:weighted-position}
 w(x)=\frac{\mathrm dP_X^{\rm tar}}{\mathrm dP_X^{\rm cal}}(x)
\end{equation}
is known. The weight $w(X_s)$ corrects for the change in the covariate
distribution between the two phases.

For within-group comparisons, use the panel
\[
 B_{t,w}=\{t\}\cup\{s:-T_0<s\le0,\ L_s=L_t\}.
\]
For each other group $k$ represented in calibration, draw $I_{k,t}$
uniformly from $\{s:-T_0<s\le0,\ L_s=k\}$, independently across groups
and of the observation values. Let $B_{t,b}$ consist of $t$ and these draws.
Use the fixed threshold scores $\widehat V_s=V(X_s,c(X_s))$, with ties
broken as in Appendix~A.1. For prespecified $q_\ell\in(0,1]$, define
\[
 \widehat V_{t,\ell}^{\rm cut}
 =
 \min\left\{
 \widehat V_s : s\in B_{t,\ell},\quad
 \sum_{u\in B_{t,\ell}}\ind\{\widehat V_u\le\widehat V_s\}
 \ge \left\lceil q_\ell |B_{t,\ell}|\right\rceil
 \right\},
 \qquad \ell\in\{w,b\}.
\]
We compute the following e-values.
\begin{equation}\label{eq:weighted-e}
\begin{aligned}
 E_{t,\ell}^{\rm shift}
 &=
 \frac{
 \bigl[\sum_{s\in B_{t,\ell}}w(X_s)\bigr]
 \ind\{\widehat V_t\le\widehat V_{t,\ell}^{\rm cut}\}
 }{
 w(X_t)+
 \sum_{s\in B_{t,\ell}\setminus\{t\}}
 w(X_s)(1-\theta_s)
 \ind\{\widehat V_s\le\widehat V_{t,\ell}^{\rm cut}\}
 },
 \qquad \ell\in\{w,b\},\\
 E_t^{\rm shift}
 &=a_tE_{t,w}^{\rm shift}+(1-a_t)E_{t,b}^{\rm shift}.
\end{aligned}
\end{equation}
Set the corresponding component to zero when its calibration set is
empty. 

\begin{proposition}[Evidence under calibration-to-online covariate shift]
\label{thm:shift}
Under \eqref{eq:phase-shift-model} and for deterministic $a_t\in[0,1]$,
\[
 \E[(1-\theta_t)E_t^{\rm shift}]\le1.
\]
Consequently, online e-BH controls FDR at every fixed time.
Averaging over the representative draws preserves the same guarantee.
\end{proposition}

\subsection{Selection under Effect-related Responses}
In applications such as treatment allocation, selection may be based on a response summarizing treatment benefit rather than the observed outcome alone. Following the treatment-effect motivation in Lee and Ren~\citep{lee2025}, we consider selecting candidates whose effect-related response $Q_t$ exceeds $c(X_t)$ before $Q_t$ is observed, where $Q_i$ is a specified response that can be evaluated from labeled calibration data or subsequently validated.

For a concrete example, let $O_i=(X_i,L_i,A_i,Y_i)$, where $A_i\in\{0,1\}$ indicates treatment and $Y_i$ is the observed outcome. Let $\widehat\mu_a(x,l)$ estimate $\E[Y_i\mid X_i=x,L_i=l,A_i=a]$ for $a\in\{0,1\}$, and let $\widehat e(x,l)$ estimate $\Prob(A_i=1\mid X_i=x,L_i=l)$. A commonly used effect-related response is the doubly robust pseudo-outcome~\citep{robins1994}
\[
 Q_i=\widehat\mu_1(X_i,L_i)-\widehat\mu_0(X_i,L_i)
 +\frac{A_i(Y_i-\widehat\mu_1(X_i,L_i))}{\widehat e(X_i,L_i)}
 -\frac{(1-A_i)(Y_i-\widehat\mu_0(X_i,L_i))}{1-\widehat e(X_i,L_i)},
\]
where $0<\widehat e<1$ and all nuisance fits are fixed from independent training. Cross-fitting is also admissible when the resulting joint array satisfies the invariance stated below.
The effect label is ${\theta_i^Q=\ind\{Q_i\ge c(X_i)\}}$. For the chosen score map $V^Q$, define the observed score $V_i^Q=V^Q(X_i,Q_i)$ and the threshold score $\widehat V_i^Q={V^Q(X_i,c(X_i))}$. The calibration null pool for group $k$ is $\calN_k^{Q,0}=\{i\le0:L_i=k,\theta_i^Q=0\}$.
The score $V^Q(x,q)$ is nondecreasing in $q$; one example is $q-\widehat\tau(x)$, with $\widehat\tau$ a fixed predictor of $Q$. OHS uses the calibration responses $Q_i$ for calibration scores and only $X_t$ for the candidate threshold score. The FDR guarantee below is defined for the labels $\theta_i^Q$, including when $Q_i$ is a pseudo-outcome.

For this extension, assume that the complete array $\{(X_{k,j},Q_{k,j})\}$ satisfies Assumption~\ref{ass:he}, and its schedule satisfies Assumption~\ref{ass:flow}. These conditions apply to the constructed effect response, not merely to the original raw outcome. Use the original OHS formulas with $(\theta,V,\widehat V,\calN)$ replaced by $(\theta^Q,V^Q,\widehat V^Q,\calN^{Q,0})$, including the between-group rank gate and deterministic weight.

\begin{proposition}[Effect-label selection]\label{prop:effect}
Under these assumptions and score monotonicity, the resulting $E_t^Q$ satisfies $\E\!\left[E_t^Q(1-\theta_t^Q)\right]\le1$, and online e-BH controls
\begin{equation}\label{eq:effect-fdr}
 \FDR^Q(t)=\E\!\left[\frac{\sum_{i\in\calR_t}(1-\theta_i^Q)}{|\calR_t|\vee1}\right]\le\alpha.
\end{equation}
\end{proposition}

\subsection{Selection Under New Group Arrivals}
Groups absent from calibration are screened from their first arrival.
When candidates arrive in a new complete group, process them one at a time in a prespecified or independently randomized order.

Let $\rho_s$ be the candidate index after whose decision the label
$\theta_s$ is released. Set $\rho_s=0$ for calibration observations.
For online candidates, $\rho_s=s$ gives immediate feedback, the last
index of the batch gives batch-end feedback, and $\rho_s=\infty$ gives
no feedback. Other scheduled delays may vary across candidates.
At decision $t$, use
\begin{equation}\label{eq:cold-pools}
 H_{k,t}=\{s:-T_0<s<t,\ L_s=k,\ \rho_s<t\},
 \qquad
 \calN_{k,t}=\{s\in H_{k,t}:\theta_s=0\}.
\end{equation}

Use the uncentered constructions with \eqref{eq:cold-pools} in
\eqref{eq:mixture} and \eqref{eq:feedback-average}, replacing $a_t$ by
\begin{equation}\label{eq:cold-weight}
 \widetilde a_t=a_t\ind\{H_{L_t,t}\ne\varnothing\}.
\end{equation}
The first candidate from a new group uses only between-group evidence.
Once a null label is released, that observation can serve as a reference
for subsequent candidates, including candidates in the same batch.

\begin{proposition}[New groups and scheduled feedback]\label{prop:cold}
Suppose Assumptions~\ref{ass:he} and~\ref{ass:flow} hold for the fixed
population of $K$ groups. The schedule, including batch boundaries,
processing order and release indices, is independent of the complete data
array. OHS under score monotonicity and OHSF with scheduled feedback
control $\FDR(t)\le\alpha$ at every fixed candidate index $t$.
\end{proposition}

\section{Simulation Studies}\label{sec:simulation}
\subsection{Basic Setup}\label{sec:sim-design}

\paragraph*{Data generation.}
We independently draw group offsets $B_k\sim N(0,1.5^2)$, signals $\theta_{k,j}\sim\operatorname{Bernoulli}(0.2)$ and feature noises $\varepsilon_{k,j,l}$. Set $Y_{k,j}=\theta_{k,j}$ and ${c(X)=1/2}$. The first three features are
\begin{equation}\label{eq:sim-model}
\begin{aligned}
 X_{k,j,1}=B_k-\Delta\theta_{k,j}+\varepsilon_{k,j,1},\quad
 X_{k,j,2}=B_k+\varepsilon_{k,j,2},\quad
 X_{k,j,3}=B_k/2+\varepsilon_{k,j,3}.
\end{aligned}
\end{equation}
\noindent The last three features are $X_{k,j,l}=\varepsilon_{k,j,l}$ for $l=4,5,6$. Noise variances are $1.5^2$ for the third coordinate and $1$ otherwise, with zero-mean Gaussian noises. Allocate $T_0=20{,}000$ calibration points equally across $K=40$ or $400$ groups and reveal $T=100$ candidates individually in fixed round-robin order. Use $\Delta\in\{3,4.5,6\}$ for $K=40$ and $\{4.5,6,8\}$ for $K=400$. Each of the 500 replications uses a fresh independent training sample of 200 groups with 50 observations each.

\paragraph*{Methods and evaluation.}
Fit an $L_2$-regularized logistic regression ($C=1$) to standardized training features, without group identifiers or latent offsets; use the negative fitted signal probability as a common fixed score. Methods share the data and paired randomization within each replication. Use $\alpha\in\{0.1,0.2,0.3\}$, $\gamma_t=0.05(0.95)^{t-1}$ without renormalization, $q_w=0.0025$, $q_b=1/K$, and \eqref{eq:weight}. Compare Within with OHS-S, OHS-R, OHS-C, OHS-RU and OHS-CU ($u=10$ for centered versions). The suffix C denotes prefix-based centering, U denotes uniform randomization of the selection rule, and CU combines both. For immediate decisions, compare OHSF with OHSF-fixed, H-e FDR and Within-p. Ungrouped online e-BH and the supplementary pooled BH comparison use pooled calibrations. See Appendix~\ref{app:comparison-methods} for the comparison methods. At each $t$,
\[
 \operatorname{Power}(t)=
 \frac{\sum_{0\le i\le t}\ind\{i\in\mathcal R_t,\theta_i=1\}}
 {1 \vee \sum_{0\le i\le t}\theta_i} \qquad
 \operatorname{FDP}(t)=
 \frac{\sum_{0\le i\le t}\ind\{i\in\mathcal R_t,\theta_i=0\}}
 {|\mathcal R_t|\vee1}.
\]
\noindent Here index $0$ is a dummy with $\theta_0=0$ and $0\notin\mathcal R_t$. Report the replicate means of power and FDP, the latter estimating FDR. Error bars show pointwise 95\% Monte Carlo intervals.

\subsection{Main Results}

\begin{figure}[!htbp]\centering
\includegraphics[width=116mm]{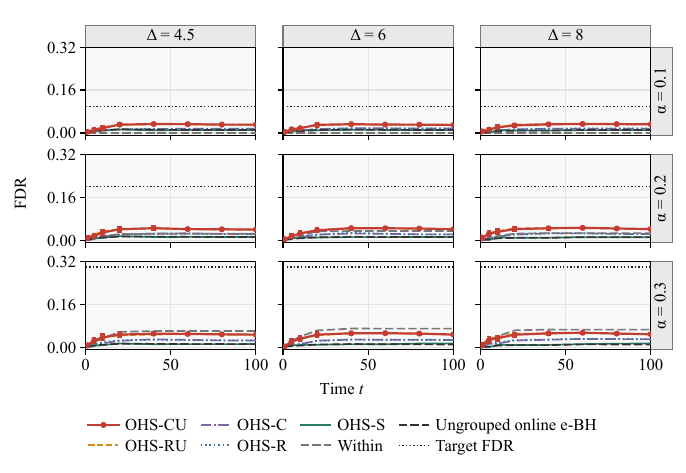}
\caption{FDR for the OHS simulation at \(K=400\). Columns vary \(\Delta\) and rows vary \(\alpha\). Error bars give pointwise 95\% Monte Carlo intervals for OHS-CU. Dotted horizontal lines show target FDR. }\label{fig:many-fdr}
\end{figure}

\begin{figure}[!htbp]\centering
\includegraphics[width=116mm]{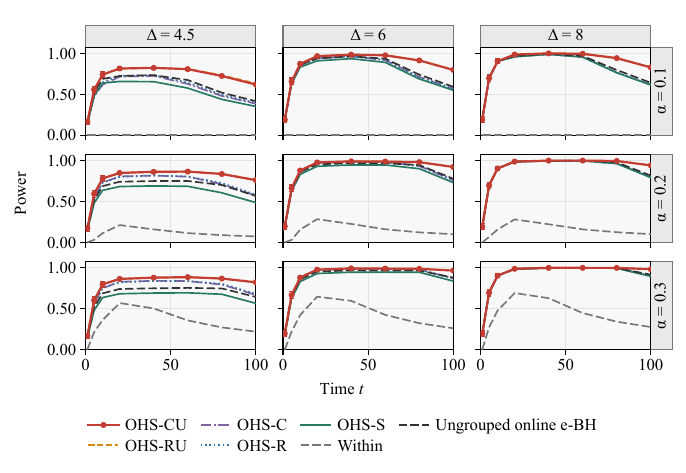}
\caption{Power for the OHS simulation at \(K=400\). Columns vary \(\Delta\) and rows vary \(\alpha\). Error bars give pointwise 95\% Monte Carlo intervals for OHS-CU. }\label{fig:many-power}
\end{figure}

Figure~\ref{fig:many-fdr} shows the empirical FDR of OHS and its baselines over 100 arrivals in the $K=400$ setting, with columns for signal strength and rows for the target FDR. Figure~\ref{fig:many-power} shows power for the same OHS comparison. All plotted empirical FDR curves remain below their nominal levels. OHS-CU and OHS-RU track each other closely and generally have the highest power. The main advantage of our method is that OHS can draw on other groups when the candidate's own calibration pool is too small to produce strong evidence. In these panels, the randomized OHS variants also improve on their unrandomized counterparts.

\begin{figure}[!htbp]\centering
\includegraphics[width=116mm]{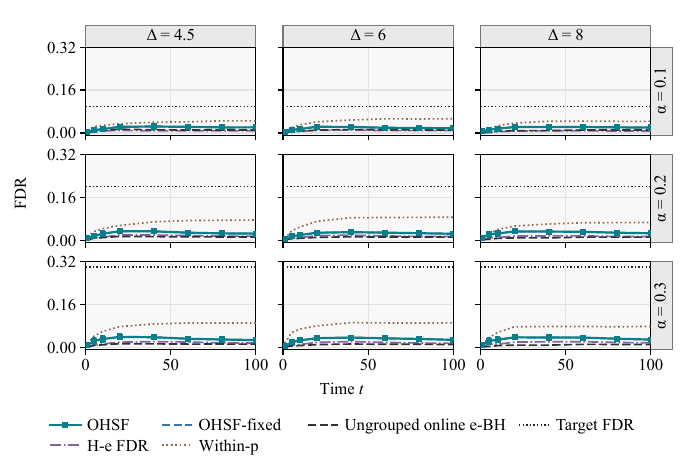}
\caption{FDR for the feedback simulation at \(K=400\). Columns vary \(\Delta\) and rows vary \(\alpha\). Error bars give pointwise 95\% Monte Carlo intervals for OHSF. Dotted horizontal lines show target FDR.}\label{fig:gaif-h-fdr}
\end{figure}

\begin{figure}[!htbp]\centering
\includegraphics[width=116mm]{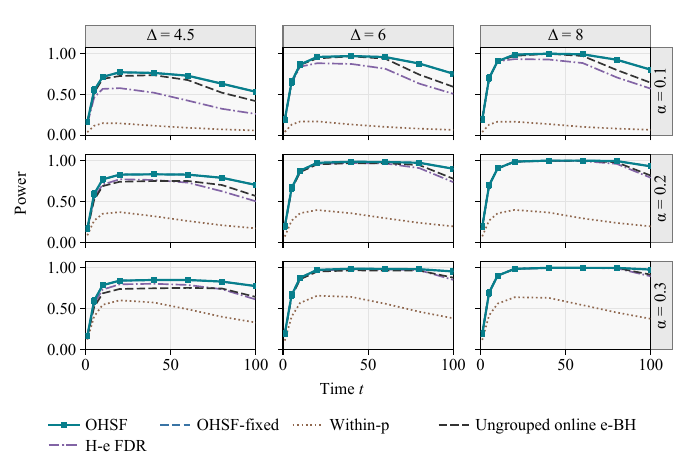}
\caption{Power for the feedback simulation at \(K=400\). Columns vary \(\Delta\) and rows vary \(\alpha\). Error bars give pointwise 95\% Monte Carlo intervals for OHSF. }\label{fig:gaif-safe}
\end{figure}

Figure~\ref{fig:gaif-h-fdr} shows the empirical FDR for OHSF and its baselines in the same $K=400$ design. All the plotted methods stay below the target levels under this setting. Figure~\ref{fig:gaif-safe} shows power for the same feedback comparison. Within-p generally has the highest empirical FDR among the feedback rules shown, whereas the OHSF and OHSF-fixed curves nearly overlap. In this experiment, OHSF uses the revealed nulls without an appreciable increase in empirical FDR relative to the fixed-pool version. OHSF and OHSF-fixed achieve similar power and outperform H-e FDR and Within-p in the displayed panels. The close agreement between OHSF and OHSF-fixed indicates that the power advantage over these baselines is already present with fixed calibration pools. Our method is very stable, and adding the revealed nulls makes little difference over the first 100 arrivals.

Taken together, Figures~\ref{fig:many-fdr}--\ref{fig:gaif-safe} show where the proposed methods are most useful in the $K=400$ design. A supplementary simulation in Appendix~\ref{app:stress} further illustrates the importance of respecting group structure: among the methods compared, only OHS keeps empirical FDR below the nominal level while retaining high power, whereas pooled BH and ungrouped online e-BH both exceed the FDR target. With the calibration sample spread across many groups, OHS can recover more true signals than Within by supplementing the same-group comparison with evidence from other groups. When a decision must be made immediately, OHSF also recovers more true signals than the within-group feedback baseline while keeping empirical FDR below the nominal level. The distinction matters: the simulations show an advantage for the selection rules that combine the two levels of evidence, and it proves that OHS and OHSF work better than other existing conformal methods under the context of grouped data. 

\section{Real Data Applications}\label{sec:case}
Section~\ref{sec:prsa-case} uses the Beijing air-quality data to screen for high PM$_{2.5}$ concentrations across monitoring stations. Section~\ref{sec:kiba-case} uses KIBA to screen drug--target pairs for high binding affinity, with groups separated by different targets.

\subsection{Beijing Air-quality Screening}\label{sec:prsa-case}
The Beijing Multi-Site Air Quality dataset (PRSA) \citep{chen2017} contains hourly pollutant and weather measurements from 12 monitoring stations. Each station is a group. The task is to identify observations with PM$_{2.5}$ concentrations at or above $15\,\mu\mathrm{g}/\mathrm{m}^{3}$ before the target measurement is available. The features include weather variables, station, cyclic time, and contemporaneous PM$_{10}$, SO$_2$, NO$_2$, CO and O$_3$, which are treated as available at screening. We use 85,667 observations from March 2013 to February 2014 for training and 21,971 complete cases sampled every eight hours from March 2014 to November 2015 for development and confirmation.

We score candidates by the negative fitted probability from a gradient-boosted classifier trained on $\ind\{Y\ge15\}$. In both applications, post-training observations in each group are split into 40\% development and 60\% confirmation pools. We run 600 paired confirmation repetitions at $\alpha=0.05$. In each repetition, we permute observations within eligible groups, truncate to a common length, randomly relabel the groups, and reveal candidates one at a time in round-robin order. Thus, the evaluation does not require exchangeability in chronological order. For PRSA, we use $m=665$ calibration observations per station, $T=160$, $a_t=0.95$, and spending weights proportional to $(t+4)^{-1}$. The within-group rank fraction is $0.01$ for the first 20 arrivals and $0.03$ thereafter. We compare OHS-RU and OHSF with OHSF-fixed, Within-U, Within-p, Alpha-spending, Pooled BH, and ungrouped online e-BH. Appendix~\ref{app:case-new} gives the within-group normalization used in both applications.

\begin{figure}[!htbp]\centering
\includegraphics[width=116mm]{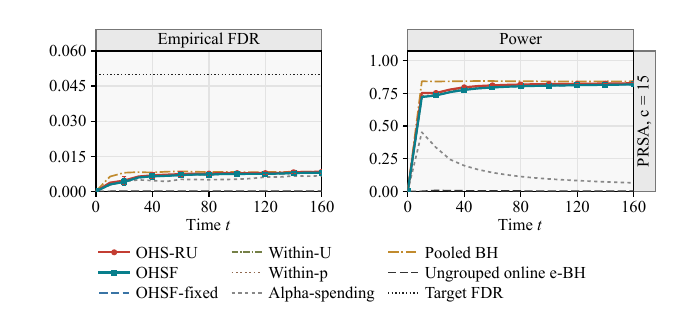}
\caption{PRSA: empirical FDR and power at $\alpha=0.05$ over 600 paired confirmation repetitions. Error bars are pointwise 95\% Monte Carlo intervals for OHS-RU and OHSF. The dotted line is the FDR target.}\label{fig:sync-prsa-Power}
\end{figure}
Figure~\ref{fig:sync-prsa-Power} shows that OHS-RU and OHSF reach power close to their within-group counterparts and have empirical FDR below $0.05$ throughout the displayed horizon. 

\subsection{Drug--target Affinity Screening in KIBA}\label{sec:kiba-case}
KIBA contains 118,254 affinity measurements for pairs drawn from 2,111 drugs and 229 kinase targets \citep{tang2014,ozturk2018}. The KIBA score combines information from different types of binding and inhibition assays on a common scale. We group drug--target pairs by target and screen for pairs with a KIBA score at or above $c=10.5$. Each arriving candidate is a pair whose measured affinity is withheld at the time of screening.

We fit a rank-48 regularized matrix factorization on 40\% of the observations for each target and use the negative predicted affinities as selection scores. We split the remaining observations into development and confirmation pools and use the same randomized revelation protocol and comparisons as in Section~\ref{sec:prsa-case}. We retain the 85 groups with at least 163 observations in each split, use $m=155$ calibration observations per target and $T=100$, and set the within-group rank fraction to $0.005$ and $a_t=0.98$. Spending weights are proportional to $(t+4)^{-1.3}$; there is no centering. Further training and configuration details are in Appendix~\ref{app:case-new}.

\begin{figure}[!htbp]\centering
\includegraphics[width=116mm]{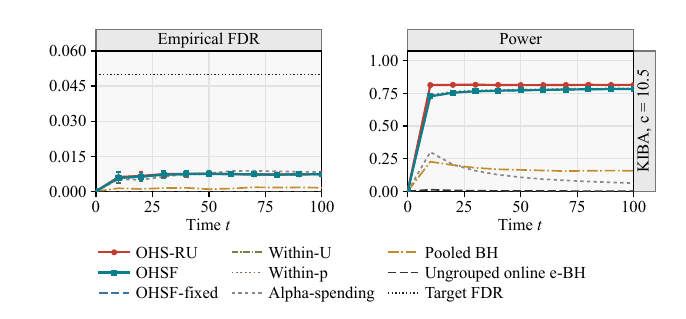}
\caption{KIBA at $c=10.5$: empirical FDR and power at $\alpha=0.05$ over 600 paired confirmation repetitions. Error bars and the dotted line have the same meanings as in Figure~\ref{fig:sync-prsa-Power}.}\label{fig:kiba-main}
\end{figure}
In Figure~\ref{fig:kiba-main}, OHS-RU and OHSF have higher terminal power than either pooled baseline, with empirical FDR below $0.05$ over the displayed horizon. Their power is close to the matched within-group methods. Appendix~\ref{app:case-new} reports the stricter threshold $c=11$.

\section{Discussion}\label{sec:conclusion}
{
This paper develops an online selection framework for selecting candidates whose unobserved responses exceed certain thresholds from hierarchical streaming data.
In the absence of response feedback, OHS constructs within-group and between-group e-values by leveraging rank evidence, and combines them through deterministic weights. By leveraging the online e-BH procedure, OHS provides anytime-valid finite-sample FDR control under the given hierarchical exchangeability and arrival assumptions. For the setting where the selection must be made immediately and the ground truth is revealed at the next time, OHSF provides a powerful selection framework. These frameworks are also extended to the broader settings, including data with covariate shift, selection under effect-related responses, and selection for unseen groups.
Simulations and real-data applications show that the proposed frameworks always control FDR and maintain powerful.

Future work could extend the feedback mechanism to settings in which labels are revealed after a delay or only for selected candidates. Another direction is to develop data-driven weights to combine within-group and between-group evidence while preserving finite-sample validity.
}

\bibliographystyle{plainnat}
\nocite{*}
\bibliography{references}

@article{gablenz2025catch,
  title={Catch me if you can: signal localization with knockoff e-values},
  author={Gablenz, Paula and Sabatti, Chiara},
  journal={J R Stat Soc Ser B Stat Methodol},
  volume={87},
  number={1},
  pages={56--73},
  year={2025},
  publisher={Oxford University Press, UK}
}

@inproceedings{
yu2026anytime,
title={Anytime Safe {PAC} Efficient Reasoning},
author={Yu, C. and Zeng, H. and Zhu, Y. and Huang, J. and Zeng, H. and Jing, B.},
booktitle={Forty-third International Conference on Machine Learning},
year={2026}}

@article{yu2024generalized,
  title={A generalized e-value feature detection method with FDR control at multiple resolutions},
  author={Yu, C. and Ming, R. and Xiao, M. and Wang, Z. and Jing, B.},
  journal={arXiv preprint arXiv:2409.17039},
  year={2024}
}

@article{lu2026,
  title={Feedback-enhanced online multiple testing with applications to conformal selection},
  author={Lu, Lin and Huo, Yuyang and Ren, Haojie and Wang, Zhaojun and Zou, Changliang},
  journal={J Mach Learn Res},
  volume={27},
  number={186},
  pages={1--86},
  year={2026}
}

@article{jin2026model,
  title={Model-free selective inference under covariate shift via weighted conformal p-values},
  author={Jin, Ying and Cand{\`e}s, Emmanuel J},
  journal={Biometrika},
  volume={113},
  number={1},
  pages={asaf066},
  year={2026},
  publisher={Oxford University Press}
}

@article{gui2024conformal,
  title={Conformal alignment: Knowing when to trust foundation models with guarantees},
  author={Gui, Yu and Jin, Ying and Ren, Zhimei},
  journal={Adv Neural Inf Process Syst},
  volume={37},
  pages={73884--73919},
  year={2024}
}

@article{yu2026robust,
  title={Robust Conformalized Selection with Noisy Responses},
  author={Yu, C. and Wei, H. and Jing, B.},
  journal={arXiv preprint arXiv:2607.22985},
  year={2026}
}

@article{price2025probabilistic,
  title={Probabilistic weather forecasting with machine learning},
  author={Price, Ilan and Sanchez-Gonzalez, Alvaro and Alet, Ferran and Andersson, Tom R and El-Kadi, Andrew and Masters, Dominic and Ewalds, Timo and Stott, Jacklynn and Mohamed, Shakir and Battaglia, Peter and others},
  journal={Nature},
  volume={637},
  number={8044},
  pages={84--90},
  year={2025},
  publisher={Nature Publishing Group UK, London}
}

@article{watson2023novo,
  title={De novo design of protein structure and function with RFdiffusion},
  author={Watson, Joseph L and Juergens, David and Bennett, Nathaniel R and Trippe, Brian L and Yim, Jason and Eisenach, Helen E and Ahern, Woody and Borst, Andrew J and Ragotte, Robert J and Milles, Lukas F and others},
  journal={Nature},
  volume={620},
  number={7976},
  pages={1089--1100},
  year={2023},
  publisher={Nature Publishing Group UK, London}
}

@article{dara2022machine,
  title={Machine learning in drug discovery: a review},
  author={Dara, Suresh and Dhamercherla, Swetha and Jadav, Surender Singh and Babu, CH Madhu and Ahsan, Mohamed Jawed},
  journal={Artif Intell Rev},
  volume={55},
  number={3},
  pages={1947--1999},
  year={2022},
  publisher={Springer}
}

@article{
stutz2023conformal,
title={Conformal prediction under ambiguous ground truth},
author={David Stutz and Abhijit Guha Roy and Tatiana Matejovicova and Patricia Strachan and Ali Taylan Cemgil and Arnaud Doucet},
journal={Trans Mach Learn Res},
issn={2835-8856},
year={2023},
url={https://openreview.net/forum?id=CAd6V2qXxc},
note={}
}

@article{su2021model,
  title={Model-assisted analyses of cluster-randomized experiments},
  author={Su, Fangzhou and Ding, Peng},
  journal={J R Stat Soc Ser B Stat Methodol},
  volume={83},
  number={5},
  pages={994--1015},
  year={2021},
  publisher={Oxford University Press}
}

@article{angelopoulos2023,
  author = {Angelopoulos, A. N. and Bates, S.},
  title = {Conformal prediction: a gentle introduction},
  journal = {Found Trends Mach Learn},
  volume = {16},
  number = {4},
  pages = {494--591},
  year = {2023},
  doi = {10.1561/2200000101}
}

@inproceedings{bai2025,
  author = {Bai, T. and Zhao, Y. and Yu, X. and Yang, A. Y.},
  title = {Multivariate conformal selection},
  booktitle = {Proceedings of the 42nd International Conference on Machine Learning},
  series = {Proceedings of Machine Learning Research},
  volume = {267},
  pages = {2535--2559},
  year = {2025}
}

@article{benjamini1995,
  author = {Benjamini, Y. and Hochberg, Y.},
  title = {Controlling the false discovery rate: a practical and powerful approach to multiple testing},
  journal = {J R Stat Soc Ser B Methodol},
  volume = {57},
  number = {1},
  pages = {289--300},
  year = {1995},
  doi = {10.1111/j.2517-6161.1995.tb02031.x}
}

@article{benjamini2001,
  author = {Benjamini, Y. and Yekutieli, D.},
  title = {The control of the false discovery rate in multiple testing under dependency},
  journal = {Ann Stat},
  volume = {29},
  number = {4},
  pages = {1165--1188},
  year = {2001},
  doi = {10.1214/aos/1013699998}
}

@misc{chen2017,
  author = {Chen, S.},
  title = {{Beijing Multi-Site Air Quality}},
  howpublished = {UCI Machine Learning Repository},
  note = {Data set},
  year = {2017},
  doi = {10.24432/C5RK5G}
}

@misc{fischer2026,
  author = {Fischer, L. and Xu, Z. and Ramdas, A.},
  title = {An online generalization of the (e-)Benjamini--Hochberg procedure},
  note = {arXiv:2407.20683v5},
  year = {2026},
  url = {https://arxiv.org/abs/2407.20683v5}
}

@article{foster2008,
  author = {Foster, D. P. and Stine, R. A.},
  title = {$\alpha$-Investing: a procedure for sequential control of expected false discoveries},
  journal = {J R Stat Soc Ser B Stat Methodol},
  volume = {70},
  number = {2},
  pages = {429--444},
  year = {2008},
  doi = {10.1111/j.1467-9868.2007.00643.x}
}

@misc{gui2025,
  author = {Gui, Y. and Jin, Y. and Nair, Y. and Ren, Z.},
  title = {{ACS}: an interactive framework for conformal selection},
  note = {arXiv:2507.15825},
  year = {2025},
  url = {https://arxiv.org/abs/2507.15825}
}

@misc{hao2025,
  author = {Hao, Q. and Liao, W. and Jing, B. and Wei, H.},
  title = {Multi-condition conformal selection},
  note = {arXiv:2510.08075},
  year = {2025},
  url = {https://arxiv.org/abs/2510.08075}
}

@article{higgins2009,
  author = {Higgins, J. P. and Thompson, S. G. and Spiegelhalter, D. J.},
  title = {A re-evaluation of random-effects meta-analysis},
  journal = {J R Stat Soc Ser A Stat Soc},
  volume = {172},
  number = {1},
  pages = {137--159},
  year = {2009},
  doi = {10.1111/j.1467-985X.2008.00552.x}
}

@article{javanmard2018,
  author = {Javanmard, A. and Montanari, A.},
  title = {Online rules for control of false discovery rate and false discovery exceedance},
  journal = {Ann Stat},
  volume = {46},
  number = {2},
  pages = {526--554},
  year = {2018},
  doi = {10.1214/17-AOS1559}
}

@article{jin2023,
  author = {Jin, Y. and Cand\`es, E. J.},
  title = {Selection by prediction with conformal p-values},
  journal = {J Mach Learn Res},
  volume = {24},
  number = {244},
  pages = {1--41},
  year = {2023}
}

@misc{lee2025,
  author = {Lee, Y. and Ren, Z.},
  title = {Selection from hierarchical data with conformal e-values},
  note = {arXiv:2501.02514},
  year = {2025},
  url = {https://arxiv.org/abs/2501.02514}
}

@article{liu2026,
  author = {Liu, K. and Xi, H. and Vong, C.-M. and Wei, H.},
  title = {Online conformal selection with accept-to-reject changes},
  journal = {Proc AAAI Conf Artif Intell},
  volume = {40},
  number = {28},
  pages = {23765--23773},
  year = {2026}
}

@article{ozturk2018,
  author = {{\"O}zt{\"u}rk, H. and {\"O}zg{\"u}r, A. and Ozkirimli, E.},
  title = {{DeepDTA}: deep drug--target binding affinity prediction},
  journal = {Bioinformatics},
  volume = {34},
  number = {17},
  pages = {i821--i829},
  year = {2018},
  doi = {10.1093/bioinformatics/bty593}
}

@inproceedings{ramdas2018,
  author = {Ramdas, A. and Zrnic, T. and Wainwright, M. J. and Jordan, M. I.},
  title = {{SAFFRON}: an adaptive algorithm for online control of the false discovery rate},
  booktitle = {Proceedings of the 35th International Conference on Machine Learning},
  series = {Proceedings of Machine Learning Research},
  volume = {80},
  pages = {4286--4294},
  year = {2018}
}

@article{robins1994,
  author = {Robins, J. M. and Rotnitzky, A. and Zhao, L. P.},
  title = {Estimation of regression coefficients when some regressors are not always observed},
  journal = {J Amer Statist Assoc},
  volume = {89},
  number = {427},
  pages = {846--866},
  year = {1994},
  doi = {10.1080/01621459.1994.10476818}
}

@article{singh2023,
  author = {Singh, R. and Sledzieski, S. and Bryson, B. and Cowen, L. and Berger, B.},
  title = {Contrastive learning in protein language space predicts interactions between drugs and protein targets},
  journal = {Proc Natl Acad Sci USA},
  volume = {120},
  number = {24},
  pages = {e2220778120},
  year = {2023},
  doi = {10.1073/pnas.2220778120}
}

@article{tang2014,
  author = {Tang, J. and Szwajda, A. and Shakyawar, S. and Xu, T. and Hintsanen, P. and Wennerberg, K. and Aittokallio, T.},
  title = {Making sense of large-scale kinase inhibitor bioactivity data sets: a comparative and integrative analysis},
  journal = {Journal of Chemical Information and Modeling},
  volume = {54},
  number = {3},
  pages = {735--743},
  year = {2014},
  doi = {10.1021/ci400709d}
}

@inproceedings{tibshirani2019,
  author = {Tibshirani, R. J. and Barber, R. F. and Cand\`es, E. J. and Ramdas, A.},
  title = {Conformal prediction under covariate shift},
  booktitle = {Advances in Neural Information Processing Systems},
  volume = {32},
  pages = {2526--2536},
  year = {2019},
  url = {https://proceedings.neurips.cc/paper/2019/hash/8fb21ee7a2207526da55a679f0332de2-Abstract.html}
}

@book{vovk2005,
  author = {Vovk, V. and Gammerman, A. and Shafer, G.},
  title = {Algorithmic Learning in a Random World},
  publisher = {Springer},
  address = {New York},
  year = {2005},
  doi = {10.1007/b106715}
}

@article{wang2022,
  author = {Wang, R. and Ramdas, A.},
  title = {False discovery rate control with e-values},
  journal = {J R Stat Soc Ser B Stat Methodol},
  volume = {84},
  number = {3},
  pages = {822--852},
  year = {2022},
  doi = {10.1111/rssb.12489}
}

@article{wang2025,
  author = {Wang, R. and Dai, R.},
  title = {Online multi-layer {FDR} control},
  journal = {Mathematics},
  volume = {13},
  number = {12},
  pages = {1937},
  year = {2025},
  doi = {10.3390/math13121937}
}

@inproceedings{xu2024,
  author = {Xu, Z. and Ramdas, A.},
  title = {Online multiple testing with e-values},
  booktitle = {Proceedings of the 27th International Conference on Artificial Intelligence and Statistics},
  series = {Proceedings of Machine Learning Research},
  volume = {238},
  pages = {3997--4005},
  year = {2024},
  url = {https://proceedings.mlr.press/v238/xu24a.html}
}

@article{yu2024,
  author = {Yu, Y. and Mai, Y. and Zheng, Y. and Shi, L.},
  title = {Assessing and mitigating batch effects in large-scale omics studies},
  journal = {Genome Biology},
  volume = {25},
  pages = {254},
  year = {2024},
  doi = {10.1186/s13059-024-03401-9}
}

@article{renbarber2024,
  author = {Ren, Z. and Barber, R. F.},
  title = {Derandomised knockoffs: leveraging e-values for false discovery rate control},
  journal = {J R Stat Soc Ser B Stat Methodol},
  volume = {86},
  number = {1},
  pages = {122--154},
  year = {2024},
  doi = {10.1093/jrsssb/qkad085}
}

@misc{xuramdas_random2023,
  author = {Xu, Z. and Ramdas, A.},
  title = {More powerful multiple testing under dependence via randomization},
  note = {arXiv:2305.11126},
  year = {2023},
  url = {https://arxiv.org/abs/2305.11126}
}

\clearpage

\appendix
\numberwithin{equation}{section}
\section{Proofs for OHS}\label{app:ohs-proofs}

\subsection{Proof of Theorem~\ref{lem:within}}
\begin{proof}
Assume ties are already broken by i.i.d. continuous marks independent of the data, suppressed in the notation and shared by $V_t$ and $\widehat V_t=V(X_t,c(X_t))$.
For $h\in\{w,b\}$, let $E_{t,h}(v)$ denote the evidence at candidate score $v$ with calibrations fixed.
Under the null, $Y_t<c(X_t)$, so monotonicity of the score gives
$V_t\le\widehat V_t$. Hence
\begin{equation}\label{eq:app-score-domination}
 (1-\theta_t)E_{t,h}(\widehat V_t)
 \le(1-\theta_t)E_{t,h}(V_t).
\end{equation}
Define
\[
 \mathcal P_t=\{i\in\{-T_0+1,\ldots,0\}:L_i=L_t\}\cup\{t\}.
\]
With the score function fixed, let $\mathcal I$ be generated by the unordered panel:
\[
 \mathcal I=\sigma\!\left((L_s)_{s=-T_0+1}^{t},
                    \{(V_i,\theta_i):i\in\mathcal P_t\}\right).
\]
If $\sum_{i\in\mathcal P_t}(1-\theta_i)\le1$, the null-weighted evidence is zero.
Otherwise, \eqref{eq:app-score-domination} and within-group exchangeability give
\begin{align}
 &\E[(1-\theta_t)E_{t,w}\mid\mathcal I]\notag\\[2pt]
 \le&
 \frac{\sum_{i\in\mathcal P_t}(1-\theta_i)}
 {\left\lceil q_w\sum_{i\in\mathcal P_t}(1-\theta_i)\right\rceil}
 \Prob\!\left(
 \theta_t=0,\;
 1+\sum_{\ell\in\mathcal P_t}(1-\theta_\ell)\ind\{V_\ell<V_t\}
 \le\left\lceil q_w\sum_{i\in\mathcal P_t}(1-\theta_i)\right\rceil
 \,\middle|\,\mathcal I\right)\notag\\[5pt]
 =&\frac{\sum_{i\in\mathcal P_t}(1-\theta_i)}
 {|\mathcal P_t|\left\lceil q_w\sum_{i\in\mathcal P_t}(1-\theta_i)\right\rceil}
 \sum_{i\in\mathcal P_t}(1-\theta_i)
 \ind\!\left\{
 1+\sum_{\ell\in\mathcal P_t}(1-\theta_\ell)\ind\{V_\ell<V_i\}
 \le\left\lceil q_w\sum_{\ell\in\mathcal P_t}(1-\theta_\ell)\right\rceil
 \right\}\notag\\[5pt]
 =&\frac{\sum_{i\in\mathcal P_t}(1-\theta_i)}
 {|\mathcal P_t|\left\lceil q_w\sum_{i\in\mathcal P_t}(1-\theta_i)\right\rceil}
 \left\lceil q_w\sum_{i\in\mathcal P_t}(1-\theta_i)\right\rceil\notag\\[3pt]
 =&\frac{\sum_{i\in\mathcal P_t}(1-\theta_i)}{|\mathcal P_t|}\le1.\label{eq:within-proof}
\end{align}
Taking expectations completes the proof:
\[
 \E[(1-\theta_t)E_{t,w}]
 =\E\!\left[\E[(1-\theta_t)E_{t,w}\mid\mathcal I]\right]\le1.
\]
\end{proof}

\subsection{Proof of Theorem~\ref{thm:between}}
\begin{proof}
Fix the score function and arrival schedule, and use the tie-breaking convention of Appendix~A.1.
Set
\[
 B=\bigl|\{L_t\}\cup\{L_s:-T_0+1\le s\le0\}\bigr|.
\]
Condition on the unordered blocks:
\[
\begin{aligned}
 \mathcal B=\sigma\!\Bigl(
 (L_s)_{s=-T_0+1}^{t},
 \Bigl\{\{(V_{k,j},\theta_{k,j}):1\le j\le n_k\}
 :k\in\{L_t\}\cup\{L_s:-T_0+1\le s\le0\}\Bigr\}\Bigr).
\end{aligned}
\]
Index these blocks and their observations arbitrarily. Independently draw
\[
 j_i\mid\mathcal B\ \stackrel{\mathrm{ind}}{\sim}\operatorname{Unif}\{1,\ldots,n_i\},\quad
 A_i=1-\theta_{i,j_i},\quad 1\le i\le B.
\]
For $1\le h\le n_i$, within-group exchangeability gives
\begin{equation}\label{eq:availability}
 1-\frac{\binom{\sum_{j=1}^{n_i}\theta_{i,j}}{h}}{\binom{n_i}{h}}
 \ge\frac1{n_i}\sum_{j=1}^{n_i}(1-\theta_{i,j})
 =\Prob(A_i=1\mid\mathcal B).
\end{equation}
Conditional on $\mathcal B$ and the assignment of blocks to roles,
the within-block orders are independent uniform permutations.
Let $p_\ell(h)$ denote the left-hand side of (A.3).
By symmetry, conditional on a nonempty pool, the selected null
representative is uniform over all null observations in its block.
For each competitor $\ell$ with $h_\ell\ge1$ calibration positions,
use independent $W_\ell\sim\mathrm{Unif}(0,1)$ to couple
\[
 A_\ell=\ind\{W_\ell\le p_\ell(1)\},
 \qquad
 C_\ell=\ind\{W_\ell\le p_\ell(h_\ell)\}\ge A_\ell,
\]
where $C_\ell$ indicates a nonempty null pool. Given $C_\ell=1$,
draw a uniform null score independently of $W_\ell$ and use it
as $V_{\ell,j_\ell}$ whenever $A_\ell=1$. For candidate block $i$,
couple its null indicator to $A_i$ and its score on $\{A_i=1\}$
to $V_{i,j_i}$. Under this coupling, on $\{A_i=1\}$, the number
of competitor scores below the candidate is at least
$
 \sum_{\ell\ne i} A_\ell
 \ind\{V_{\ell,j_\ell}<V_{i,j_i}\}.
$

Finally, conditional on $\mathcal B$, each of the $B$ blocks occupies
the candidate role with probability $1/B$ by block exchangeability.
If $B=1$, the evidence is zero.
Otherwise, by block exchangeability, \eqref{eq:app-score-domination}, \eqref{eq:availability}, and
\[
 \frac{1+|\calK_t|}{\min\{\lceil q_bK\rceil,1+|\calK_t|\}}
 \le\frac{B}{\min\{\lceil q_bK\rceil,B\}},
\]
we have
\begin{equation}\label{eq:telescope}
\begin{aligned}
 &\E[(1-\theta_t)E_{t,b}\mid\mathcal B]\\[2pt]
 \le&\frac1{\min\{\lceil q_bK\rceil,B\}}\sum_{i=1}^{B}
 \Prob\!\left(A_i=1,\;
 \sum_{\ell\ne i}A_\ell\ind\{V_{\ell,j_\ell}<V_{i,j_i}\}
 <\min\{\lceil q_bK\rceil,B\}
 \,\middle|\,\mathcal B\right)\\[5pt]
 =&\frac1{\min\{\lceil q_bK\rceil,B\}}
 \E\!\left[\sum_{i=1}^{B}A_i\ind\!\left\{
 \sum_{\ell\ne i}A_\ell\ind\{V_{\ell,j_\ell}<V_{i,j_i}\}
 <\min\{\lceil q_bK\rceil,B\}\right\}
 \,\middle|\,\mathcal B\right]\\[5pt]
 =&\frac1{\min\{\lceil q_bK\rceil,B\}}
 \E\!\left[\min\!\left\{\lceil q_bK\rceil,\sum_{i=1}^{B}A_i\right\}
 \,\middle|\,\mathcal B\right]\le1.
\end{aligned}
\end{equation}
Hence,
\[
 \E[(1-\theta_t)E_{t,b}]
 =\E\!\left[\E[(1-\theta_t)E_{t,b}\mid\mathcal B]\right]\le1.
\]
\end{proof}

\subsection{Proof of Theorem~\ref{thm:ohs}}
\begin{proof}
By Theorems~\ref{lem:within} and~\ref{thm:between},
\[
 \E\!\left[E_t(1-\theta_t)\right]
 \le a_t+(1-a_t)=1.
\]
Suppose $k_t^*>0$. By construction, $|\calR_t|\ge k_t^*$.
If $|\calR_t|>k_t^*$, since
$1/|\calR_t|<1/k_t^*$, every $i\in\calR_t$ would also satisfy
$\alpha\gamma_iE_i\ge 1/|\calR_t|$. Thus, $|\calR_t|$ would be a feasible
step-up count, contradicting the maximality of $k_t^*$.
It follows that $|\calR_t|=k_t^*$.
For each selected index, the selection threshold now gives
$1/|\calR_t|\le\alpha\gamma_iE_i$. Summing over the selected
nulls yields
\begin{equation}\label{eq:ohs-proof}
 \frac{\sum_{j\in\calR_{t}}(1-\theta_j)}{|\calR_{t}|\vee1}
 =\sum_{i\le t}(1-\theta_i)
   \frac{\ind\{i\in\calR_t\}}{|\calR_{t}|\vee1}
 \le\alpha\sum_{i\le t}\gamma_i E_i(1-\theta_i).
\end{equation}
By the upper result, we can finish this proof
\[
 \FDR(t)\le\alpha\sum_{i\le t}\gamma_i\E[E_i(1-\theta_i)]
 \le\alpha\sum_{i\le t}\gamma_i\le\alpha.
\]
For a specified full sequence and any almost surely finite stopping time \(\tau\), the same pathwise bound gives
\[
 \E\!\left[\frac{\sum_{j\in\calR_{\tau}}(1-\theta_j)}{|\calR_{\tau}|\vee1}\right]
 \le\alpha\sum_{i\ge1}\gamma_i\E[E_i(1-\theta_i)]\le\alpha.
\]
This proof is motivated by Tonelli \citep{wang2022,fischer2026}.
\end{proof}

\subsection{Proof of Remark~\ref{prop:average}}
\begin{proof}
Conditional on $\mathcal G_t$, the representative draws do not depend on the candidate's null status $\theta_t$. Thus the same average is obtained whether or not we also condition on $\theta_t$:
\[
 \E[E_{t,b}\mid\mathcal G_t,\theta_t]
 =\E[E_{t,b}\mid\mathcal G_t]
 =\overline E_{t,b}.
\]
Conditioning on $\theta_t$ here is only a proof step; the average in \eqref{eq:rb} uses $\mathcal G_t$ alone. By iterated expectation and Theorem~\ref{thm:between},
\begin{equation}\label{eq:rb-proof}
 \E\!\left[\overline E_{t,b}(1-\theta_t)\right]
 =\E\!\left[E_{t,b}(1-\theta_t)\right]\le1.
\end{equation}
Apply deterministic mixing and Theorem~\ref{thm:ohs}.
\end{proof}

\subsection{Proof of Remark~\ref{prop:center}}
\begin{proof}
Let $\Phi$ apply the group-symmetric prefix fit and retain the transformed suffixes. For a suffix permutation $\pi$ and a complete-block permutation $\sigma$,
\[
 \Phi(\pi\calZ)=\pi\Phi(\calZ),\qquad
 \Phi(\sigma\calZ)=\sigma\Phi(\calZ).
\]
Indeed, $\pi$ fixes the fitting prefixes and $\sigma$ moves them with their blocks. Assumption~\ref{ass:he} implies
\[
 \Phi(\calZ)\Ed\pi\Phi(\calZ),\qquad
 \Phi(\calZ)\Ed\sigma\Phi(\calZ).
\]
The suffix schedule is feasible and independent of this array. Let $b_k$ and $s_k>0$ be the location and scale fitted on group $k$'s prefix, respectively ($b_k=\widehat\mu_k$ and $s_k=1$ for centering alone). Since $s_{L_t}>0$, whenever $V_t\le\widehat V_t$, we have
\[
 \frac{V_t-b_{L_t}}{s_{L_t}}
 \le \frac{\widehat V_t-b_{L_t}}{s_{L_t}}.
\]
The same comparison holds for any response-monotone adjustment. Apply Theorem~\ref{thm:between} and Remark~\ref{prop:average} to the suffixes. With the unchanged within-group evidence,
\[
 \E[(1-\theta_t)E_t]\le a_t+(1-a_t)=1.
\]
\end{proof}

\subsection{Proof of Proposition~\ref{thm:boost}}
\begin{proof}
Fix a time $t$, all unboosted evidence and labels, and all uniforms except $U_i$. Within this proof, $\calR_t$ and $k_t^*$ refer to the selected set and step-up count for the boosted inputs. For $\gamma_iE_i>0$, replace input $i$ by infinity and denote the resulting step-up count by $k_i^0\ge1$. By step-up maximality, candidate $i$ is selected precisely when
$U_i\le\alpha\gamma_i E_i k_i^0$. On this event,
$|\calR_t|=k_i^0$.
Indeed, passing at $k_i^0$ preserves feasibility there. Conversely, if $i$ is selected at $k_t^*$, it already passes at every $k>k_t^*$, so replacing it by infinity adds no feasible $k$. Since $k_i^0$ does not depend on $U_i$,
\[
 \int_0^1\frac{\ind\{i\in\calR_t\}}{|\calR_t|\vee1}\,dU_i
 =\frac{\min\{1,\alpha\gamma_iE_i k_i^0\}}{k_i^0}
 \le\alpha\gamma_iE_i.
\]
Terms with \(\gamma_iE_i=0\) vanish. Summing the preceding integral,
\[
 \begin{aligned}
 \FDR(t)
 &=\sum_{i\le t}\E\!\left[
 (1-\theta_i)\int_0^1
 \frac{\ind\{i\in\calR_t\}}{|\calR_t|\vee1}\,dU_i\right]
 \le\alpha\sum_{i\le t}\gamma_i\E[(1-\theta_i)E_i]\le\alpha .
 \end{aligned}
\]
Since $E_i/U_i\ge E_i$ for $0<U_i\le1$, we have
\[
 \calR_t^{\mathrm{unboosted}}\subseteq\calR_t^{\mathrm{boosted}}.
\]
This uses step-up monotonicity and requires no mean bound on the boosted inputs.
\end{proof}

\section{Proofs for OHSF}\label{app:gaif-proofs}
\subsection{Proof of Theorem~\ref{thm:gaif-fdr}}
\begin{proof}
Fix the score function and arrival schedule, and use the tie-breaking convention of Appendix~A.1.
Define
$
 H_{k,t}=\{i\in\{-T_0+1,\ldots,t-1\}:L_i=k\},
$ and $
\mathcal P_t=H_{L_t,t}\cup\{t\}.
$
Condition on the unordered panel:
\[
 \mathcal I=\sigma\!\left((L_s)_{s=-T_0+1}^{t},
                 \{(\widehat V_i,\theta_i):i\in\mathcal P_t\}\right).
\]
If $\sum_{i\in\mathcal P_t}(1-\theta_i)\le1$, then $(1-\theta_t)E_{t,w}=0$.
Otherwise, within-group exchangeability gives
\[
\begin{aligned}
 &\quad~\E[(1-\theta_t)E_{t,w}\mid\mathcal I]\\[2pt]
 &=\frac{\sum_{i\in\mathcal P_t}(1-\theta_i)}
 {|\mathcal P_t|\left\lceil q_w\sum_{i\in\mathcal P_t}(1-\theta_i)\right\rceil}
 \sum_{i\in\mathcal P_t}(1-\theta_i)
 \ind\!\left\{
 1+\sum_{\ell\in\mathcal P_t}(1-\theta_\ell)
                  \ind\{\widehat V_\ell<\widehat V_i\}
 \le\left\lceil q_w\sum_{\ell\in\mathcal P_t}(1-\theta_\ell)\right\rceil
 \right\}\\[5pt]
 &=\frac{\sum_{i\in\mathcal P_t}(1-\theta_i)}
 {|\mathcal P_t|\left\lceil q_w\sum_{i\in\mathcal P_t}(1-\theta_i)\right\rceil}
 \left\lceil q_w\sum_{i\in\mathcal P_t}(1-\theta_i)\right\rceil\\[3pt]
 &=\frac{\sum_{i\in\mathcal P_t}(1-\theta_i)}{|\mathcal P_t|}\le1.
\end{aligned}
\]
For the between-group component, apply \eqref{eq:availability}--\eqref{eq:telescope}
with exposure $h=|H_{k,t}|$ and scores $\widehat V_i$ on both sides, followed by
\eqref{eq:rb-proof}. Under centering, first remove the fitting prefixes and use
the adjusted suffix scores as in Remark~\ref{prop:center}. Thus,
\[
 \E[(1-\theta_t)\overline E_{t,b}]\le1.
\]
Taking expectations and applying \eqref{eq:feedback-average},
\begin{equation}\label{eq:feedback-e-valid}
\begin{aligned}
 \E[(1-\theta_t)\overline E_t]
 =a_t\E[(1-\theta_t)E_{t,w}]
   +(1-a_t)\E[(1-\theta_t)\overline E_{t,b}]
 \le a_t+(1-a_t)=1.
\end{aligned}
\end{equation}
For deterministic $u\ge0$, integrate $U_t$:
\begin{equation}\label{eq:e-to-p-proof}
 \Prob(\theta_t=0,U_t\le u\overline E_t)
 =\E[(1-\theta_t)\min\{1,u\overline E_t\}]
 \le u\E[(1-\theta_t)\overline E_t]\le u.
\end{equation}
To bound the FDR, let $\mathcal D$ denote the complete data, tie marks and fixed score information. Full feedback makes $\overline E_t$ measurable with respect to $\mathcal D$. Set $\mathcal H_{t-1}=\sigma(\mathcal D,U_1,\ldots,U_{t-1})$. Then
\[
 (\theta_t,\overline E_t,\alpha_{t,R})\text{ is }\mathcal H_{t-1}\text{-measurable},
 \qquad U_t\perp\!\!\!\perp\mathcal H_{t-1}.
\]
Irreversibility gives that as long as $\delta_{t}=1$, 
\[
 |\calR_T|\vee1\ge1+|\calR_{t-1}|.
\]
Hence
\[
 \begin{aligned}
 \E\!\left[\left.\frac{(1-\theta_t)\delta_t}{1+|\calR_{t-1}|}\right|\mathcal H_{t-1}\right]
 =\frac{(1-\theta_t)\min\{1,\alpha_{t,R}\overline E_t\}}{1+|\calR_{t-1}|}
 \le\alpha\gamma_t(1-\theta_t)\overline E_t.
 \end{aligned}
\]
Consequently,
\[
\FDR(T)\le\sum_{t\le T}\E\!\left[\frac{(1-\theta_t)\delta_t}{1+|\calR_{t-1}|}\right]
\le\alpha\sum_{t\le T}\gamma_t\E[(1-\theta_t)\overline E_t]\le\alpha.
\]

\end{proof}

\section{Proofs for the Extensions}\label{app:extension-proofs}
\subsection{Proof of Proposition~\ref{thm:shift}}
\begin{proof}
Fix the score function, arrival schedule, and representative indices.
For $\ell\in\{w,b\}$, set $B=B_{t,\ell}$. The component is zero if
$B=\{t\}$. For $|B|>1$, let $\E_0$ denote expectation under
\eqref{eq:phase-shift-model} with the candidate covariate drawn from
$P_X^{\rm cal}$. Since the conditional response law is common,
\[
 \E[(1-\theta_t)E_{t,\ell}^{\rm shift}]
 =
 \E_0[w(X_t)(1-\theta_t)E_{t,\ell}^{\rm shift}],
\]
where expectations in this calculation are conditional on the fixed quantities.
Use the cutoff in Section~4.1 and write
\[
 D=\sum_{s\in B} w(X_s)(1-\theta_s)
       \ind\{\widehat V_s\le\widehat V_{t,\ell}^{\rm cut}\}.
\]
On $\{\theta_t=0,\ \widehat V_t\le\widehat V_{t,\ell}^{\rm cut}\}$,
the denominator in \eqref{eq:weighted-e} equals $D$.
With $0/0=0$ for the weighted terms below, exchangeability under $\E_0$ gives
\[
\begin{aligned}
 \E[(1-\theta_t)E_{t,\ell}^{\rm shift}]
 &=\E_0\!\left[
   \frac{\sum_{s\in B}w(X_s)}{D}
   w(X_t)(1-\theta_t)
   \ind\{\widehat V_t\le\widehat V_{t,\ell}^{\rm cut}\}
   \right]\\
 &=\E_0\!\left[
   \frac{\sum_{s\in B}w(X_s)}{|B|}\ind\{D>0\}
   \right]
 \le \frac{1}{|B|}\sum_{s\in B}\E_0[w(X_s)]
 =1.
\end{aligned}
\]
Averaging over the representative draws and applying \eqref{eq:weighted-e},
\[
 \E[(1-\theta_t)E_t^{\rm shift}]
 =
 a_t\E[(1-\theta_t)E_{t,w}^{\rm shift}]
 +(1-a_t)\E[(1-\theta_t)E_{t,b}^{\rm shift}]
 \le 1.
\]
For the e-value averaged over the representative draws, the same bound
follows by iterated expectation conditional on the complete data and tie marks.
Apply \eqref{eq:ohs-proof} to obtain fixed-time FDR control.
\end{proof}

\subsection{Proof of Proposition~\ref{prop:effect}}
\begin{proof}
Apply the assumed hierarchical exchangeability and content-free arrival conditions to the constructed effect-response array. For $h\in\{w,b\}$, write $E_{t,h}^Q(v)$ for the effect-label evidence with candidate score $v$ and the effect-null calibrations fixed. On the effect null,
\[
\begin{aligned}
 \theta_t^Q=0&\ \Longrightarrow\ V^Q(X_t,Q_t)\le {V^Q(X_t,c(X_t))}\\
 &\ \Longrightarrow\ E_{t,h}^Q\bigl({V^Q(X_t,c(X_t))}\bigr)
 \le E_{t,h}^Q\bigl(V^Q(X_t,Q_t)\bigr),\quad h\in\{w,b\}.
\end{aligned}
\]
With calibration scores fixed, apply \eqref{eq:within-proof} and \eqref{eq:telescope} to the effect array:
\[
 \E[(1-\theta_t^Q)E_{t,h}^Q]\le1,\qquad
 \E[(1-\theta_t^Q)E_t^Q]\le a_t+(1-a_t)=1.
\]
Equation~\eqref{eq:ohs-proof} gives \eqref{eq:effect-fdr}.
\end{proof}

\subsection{Proof of Proposition~\ref{prop:cold}}
\begin{proof}
Fix the schedule. Apply the within-group rank calculation
\eqref{eq:within-proof} to $H_{L_t,t}\cup\{t\}$. In the between-group
calculation \eqref{eq:availability}--\eqref{eq:telescope}, use
exposure $|H_{k,t}|$ for each competing group with $H_{k,t}\ne\varnothing$.
For OHSF use $\widehat V$ for candidates and references and apply
\eqref{eq:rb-proof}. Conditional on the schedule, each component has
null-weighted mean at most one. Since \eqref{eq:cold-weight} is fixed
given the schedule,
\[
 \E[(1-\theta_t)E_t]\le1,
 \qquad
 \E[(1-\theta_t)\overline E_t]\le1,
\]
where the expectations are conditional on the schedule.
For OHS, \eqref{eq:ohs-proof} gives
\[
 \FDR(t)\le\alpha\sum_{i=1}^t\gamma_i\E[(1-\theta_i)E_i]\le\alpha.
\]

For OHSF, let $\mathcal D$ contain the complete data, the schedule
and the tie marks. Then $U_i$ is independent of
$\sigma(\mathcal D,U_1,\ldots,U_{i-1})$, given which
$\alpha_{i,R}$ and $\overline E_i$ are fixed. Using
$|\calR_t|\vee1\ge1+|\calR_{i-1}|$ on $\{\delta_i=1\}$ gives
\[
\begin{aligned}
 \FDR(t)
 \le \sum_{i=1}^t\E\!\left[
 \frac{(1-\theta_i)\min\{1,\alpha_{i,R}\overline E_i\}}
      {1+|\calR_{i-1}|}\right]\le\alpha\sum_{i=1}^t\gamma_i\E[(1-\theta_i)\overline E_i]
 \le\alpha .
\end{aligned}
\]
\end{proof}

\section{Additional Details and Results of Experiments}
\subsection{Comparison Methods}\label{app:comparison-methods}\label{app:baselines}
The labels in Table~\ref{tab:comparator-labels} match the simulation
legends in Figures~\ref{fig:many-fdr}--\ref{fig:gaif-safe}, the real-data
legends in Figures~\ref{fig:sync-prsa-Power} and~\ref{fig:kiba-main}, and
the supplementary figures in Appendices~\ref{app:stress} and~\ref{app:case-new}.
For real-data normalization, see Appendix~\ref{app:case-new}.

OHS uses the hierarchical subsampling idea of Lee and Ren~\citep{lee2025}
with online e-BH~\citep{fischer2026}. The U variants apply uniform
randomization~\citep{xuramdas_random2023} to the inputs of online e-BH.
OHSF and its matched decision variants use the e-LOND spending
rule~\citep{xu2024} as specified in \eqref{eq:gaif-r}.

\begin{table}[!htbp]
\centering\small
\caption{Method labels and decision rules used in the experiments.}
\label{tab:comparator-labels}
\begin{tabular}{@{}p{.29\linewidth}p{.66\linewidth}@{}}
\toprule
Figure label & Evidence and decision rule \\
\midrule
OHS-S & Sample one null representative per eligible group and apply online
e-BH to the combined e-value in \eqref{eq:mixture}. \\
OHS-R & Use the exact average in \eqref{eq:rb}, following the
e-value averaging principle of Ren and Barber~\citep{renbarber2024},
and apply online e-BH. \\
OHS-C & Apply OHS-R with the between-group centering in \eqref{eq:prefix}. \\
OHS-RU & Apply online e-BH to the OHS-R e-value divided by $U_t$. \\
OHS-CU & Apply online e-BH to the OHS-C e-value divided by $U_t$. \\
OHS-SU & Apply online e-BH to the OHS-S e-value divided by $U_t$. \\
Within & Apply online e-BH to the fixed-calibration within-group
e-value in \eqref{eq:within}. \\
Within-U & Apply online e-BH to the Within e-value divided by $U_t$. \\
OHSF & Update the reference pools by \eqref{eq:feedback-pools} and apply
the randomized decision rule in \eqref{eq:gaif-r}. \\
OHSF-fixed & Use the OHSF decision rule with $\calN_{k,t}=\calN_k$. \\
H-e FDR & Use the OHSF evidence and set $U_t=1$ in \eqref{eq:gaif-r}. \\
Within-p & Apply \eqref{eq:gaif-r} with the updated
within-group e-value in \eqref{eq:p-within} as its input. \\
Alpha-spending & Select at time $t$ if $U_t\le\alpha\gamma_t\overline E_t$,
using the OHSF evidence in \eqref{eq:feedback-average}. \\
Pooled BH & Apply BH~\citep{benjamini1995} at each displayed prefix to the pooled
p-values defined below. \\
Ungrouped online e-BH & Apply \eqref{eq:online-ebh}
to the pooled e-values in \eqref{eq:pooled}. \\
\bottomrule
\end{tabular}
\end{table}

\paragraph{Pooled calibration.}
Use $\calN^{\mathrm{pool}}=\bigcup_k\calN_k$ and
$N_t^{\mathrm{pool}}=1+|\calN^{\mathrm{pool}}|$. With the same score
comparison as in \eqref{eq:within}, define
\begin{equation}\label{eq:pooled}
\begin{aligned}
 p_t^{\mathrm{pool}}
 &=\frac{1+\sum_{i\in\calN^{\mathrm{pool}}}
                    \ind\{V_i<\widehat V_t\}}{N_t^{\mathrm{pool}}},\\
 r_t^{\mathrm{pool}}&=\lceil q_wN_t^{\mathrm{pool}}\rceil,\\
 E_t^{\mathrm{pool}}
 &=\frac{N_t^{\mathrm{pool}}}{r_t^{\mathrm{pool}}}
   \ind\{p_t^{\mathrm{pool}}\le r_t^{\mathrm{pool}}/N_t^{\mathrm{pool}}\}.
\end{aligned}
\end{equation}
Set $E_t^{\mathrm{pool}}=0$ when $\calN^{\mathrm{pool}}$ is empty.

\paragraph{Paired comparison of OHSF and H-e FDR.}
Write $\calR_t^{\rm unrand}$ for the H-e FDR selection set, starting
from $\calR_0^{\rm unrand}=\varnothing$. Both rules use the same
$\overline E_t$, and each computes its threshold from its own selection
history. If $\calR_{t-1}^{\rm unrand}\subseteq\calR_{t-1}$, the H-e FDR
threshold is at most the OHSF threshold in \eqref{eq:gaif-r}. Since
$U_t\le1$, induction gives $\calR_t^{\rm unrand}\subseteq\calR_t$.
For H-e FDR, on $\{t\in\calR_t^{\rm unrand}\}$,
\[
 \frac{1}{|\calR_T^{\rm unrand}|\vee1}
 \le \frac{1}{1+|\calR_{t-1}^{\rm unrand}|}
 \le \alpha\gamma_t\overline E_t.
\]
Summing over null selections and using the marginal bound under
Theorem~\ref{thm:gaif-fdr} gives
\[
 \FDR^{\rm unrand}(T)
 \le\alpha\sum_{t=1}^T\gamma_t\E[(1-\theta_t)\overline E_t]
 \overset{\eqref{eq:feedback-e-valid}}{\le}\alpha.
\]

\subsection{Additional Results of Simulations}\label{app:stress}
The following table uses the same fitted-score repetitions as Section~\ref{sec:simulation}. Each design has 500 repetitions, generating 200 external training groups of 50 observations each under \eqref{eq:sim-model}. In each repetition, standardize the features and fit the logistic model described in Section~\ref{sec:sim-design}, then freeze scores for calibration and selection. All levels share the fit. The total calibration size is 20,000, $T=100$, $q_w=0.0025$, the centering prefix is 10, and $\gamma_t=0.05(0.95)^{t-1}$ without renormalization.
Each repetition has separate training, evaluation and procedural streams. Table~\ref{tab:trained-supp} replaces the earlier directly generated-score results.

\begin{table}[ht]\centering\small\caption{Terminal Power/FDR with fitted scores.}\label{tab:trained-supp}\begin{tabular}{lllllll}\toprule
$K$&$\Delta$&$\alpha$&Within&OHS-C&OHS-CU&OHSF\\\midrule
400&4.5&0.1&0.000/0.000&0.385/0.016&0.618/0.031&0.530/0.021\\
400&4.5&0.2&0.073/0.024&0.572/0.023&0.761/0.040&0.702/0.025\\
400&4.5&0.3&0.220/0.062&0.669/0.026&0.823/0.048&0.777/0.028\\
400&6&0.1&0.000/0.000&0.568/0.017&0.798/0.030&0.751/0.018\\
400&6&0.2&0.100/0.034&0.759/0.022&0.922/0.041&0.900/0.024\\
400&6&0.3&0.261/0.070&0.871/0.028&0.967/0.048&0.955/0.028\\
400&8&0.1&0.000/0.000&0.615/0.016&0.828/0.033&0.800/0.021\\
400&8&0.2&0.101/0.027&0.796/0.023&0.942/0.042&0.930/0.026\\
400&8&0.3&0.276/0.067&0.901/0.030&0.983/0.049&0.979/0.029\\
40&3&0.1&0.104/0.026&0.095/0.026&0.212/0.040&0.172/0.036\\
40&3&0.2&0.188/0.034&0.179/0.033&0.284/0.042&0.255/0.034\\
40&3&0.3&0.230/0.033&0.223/0.034&0.328/0.051&0.294/0.038\\
40&4.5&0.1&0.370/0.014&0.353/0.013&0.571/0.021&0.494/0.017\\
40&4.5&0.2&0.533/0.017&0.519/0.017&0.695/0.027&0.657/0.020\\
40&4.5&0.3&0.626/0.018&0.611/0.018&0.756/0.032&0.722/0.022\\
40&6&0.1&0.527/0.018&0.510/0.018&0.743/0.021&0.675/0.019\\
40&6&0.2&0.708/0.017&0.690/0.017&0.866/0.024&0.838/0.020\\
40&6&0.3&0.806/0.017&0.792/0.018&0.920/0.030&0.902/0.020\\
\bottomrule\end{tabular}\end{table}

\paragraph{Supplementary simulation: the importance of OHS under grouped settings.}\label{app:pooled-stress}
Using ungrouped methods can also lead to FDR control failures in grouped settings. To illustrate this, we generate $K=10$ independent, equally sized group blocks. Group $k$ has a latent offset $B_k$, drawn uniformly from $\{-6,6\}$. Within each block, we independently draw labels $\theta_{k,j}\sim\operatorname{Bernoulli}(0.35)$ and errors $\varepsilon_{k,j}\sim N(0,0.6^2)$, independently of the offsets.
Set
\[
 X_{k,j}=B_k+4\theta_{k,j}+\varepsilon_{k,j},
 \qquad Y_{k,j}=\theta_{k,j},\qquad c=1/2.
\]
The groups share the same block distribution. The latent offsets shift the
null score distributions of the realized groups. The schedule, fixed before
observing the blocks, assigns 19,100 calibration observations to group~1 and
100 to each of the other nine groups. We then reveal $T=100$ candidates one
at a time in round-robin order from groups~2 through~10. 

We compare the sampled OHS procedure (OHS-S), pooled BH, and ungrouped online
e-BH at $\alpha=0.10$. The pooled methods use ranks against the same pool of
all offline nulls, ignoring group membership. Pooled BH applies BH to the
pooled rank $p$-values at each displayed prefix; ungrouped online e-BH uses
the pooled e-values in \eqref{eq:pooled}. Both e-BH procedures use
$q_w=0.01$ and $\gamma_t=1/T$ for $1\le t\le T$. OHS-S uses $q_b=0.2$
and the weight in \eqref{eq:weight}. We used 120 pilot replications to
check the design and report 800 independent paired confirmation
replications.

\begin{figure}[!htbp]
 \centering
 \includegraphics[width=116mm]{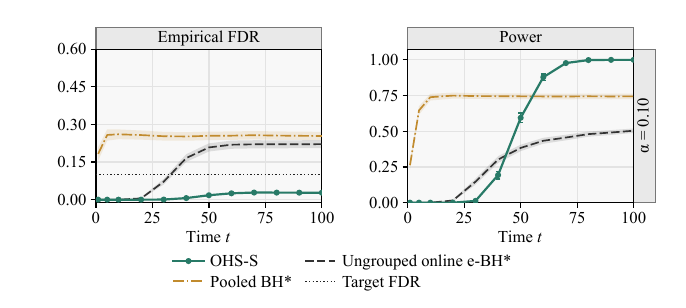}
 \caption{Empirical FDR and power over 800 paired confirmation replications. Shading gives pointwise 95\% Monte Carlo intervals and the dotted line marks the nominal FDR. Asterisks denote pooled baselines without an FDR guarantee under hierarchical exchangeability.}
 \label{fig:pooled-stress}
\end{figure}

Figure~\ref{fig:pooled-stress} shows that both pooled baselines exceed the
FDR target in this design. At $T=100$, their empirical FDRs are $0.254$
(pooled BH) and $0.221$ (ungrouped online e-BH), compared with $0.028$ for
OHS-S. The corresponding powers are $0.744$, $0.504$, and $1.000$,
respectively. In this stress design the OHS weight is close to one, so the result illustrates the value of preserving group-specific calibration when pooling is misleading, which clearly shows that to maintain FDR control, it's necessary to use OHS instead of ungrouped methods in hierarchical settings.

\begin{figure}[!htbp]\centering
\includegraphics[width=116mm]{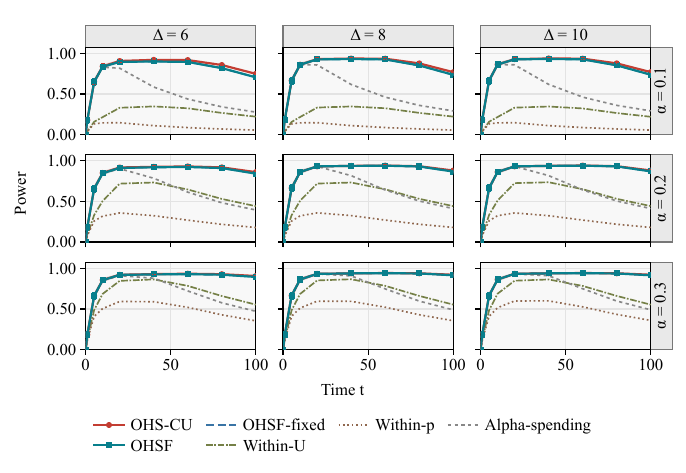}
\caption{Power with $c(X)=0.4+0.15\tanh(X_2)$ and $T=100$.
Rows correspond to $\alpha$ and columns to $\Delta$.
Curves are averages over 500 independent confirmation repetitions;
intervals for OHS-CU and OHSF are pointwise 95\% Monte Carlo intervals.
}\label{fig:supp-cx-power}
\end{figure}

\begin{figure}[!htbp]\centering
\includegraphics[width=116mm]{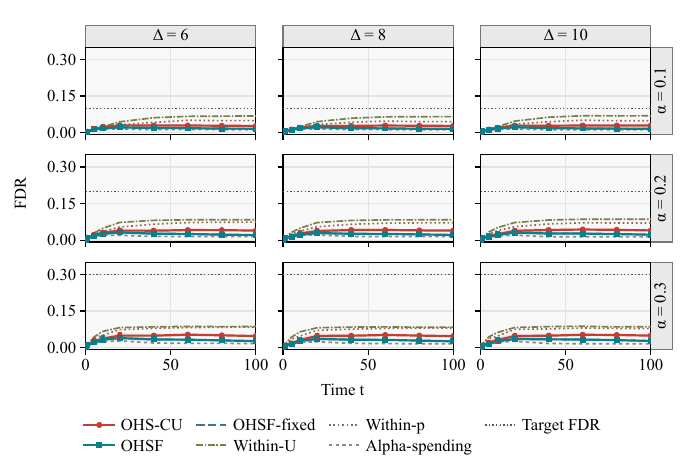}
\caption{Empirical FDR under the setting of Figure~\ref{fig:supp-cx-power}.
The dotted reference is the nominal level in each row.}\label{fig:supp-cx-fdr}
\end{figure}

\FloatBarrier
\paragraph{Supplementary simulation: a covariate-dependent threshold.}
We retain the feature model, independent logistic training, $K=400$,
$T_0=20{,}000$, $q_w=0.0025$, and the centering prefix $u=10$.
For this separate experiment, set $Y=\eta+\varepsilon$ with
$\varepsilon\sim N(0,0.15^2)$ independent of the original binary $\eta$, and use
\[
 c(X)=0.4+0.15\tanh(X_2),\qquad T=100.
\]
Fit the same standardized logistic regression to $\ind\{Y\ge c(X)\}$
and use its negative fitted probability as the common feature score.
The original spending sequence $\gamma_t=0.05(0.95)^{t-1}$ is unchanged.
We select the threshold setting on 40 development repetitions and use 500
independent confirmation repetitions at $\alpha\in\{0.1,0.2,0.3\}$ and
$\Delta\in\{4.5,6,8,10\}$. Figures~\ref{fig:supp-cx-power}--\ref{fig:supp-cx-fdr}
display the prespecified subset $\Delta\in\{6,8,10\}$.
The threshold changes approximately 1.19\% of labels relative to $c=0.4$;
comparisons across different thresholds concern different selection tasks.
Across the nine displayed settings, the minimum mean power of OHS-CU and
OHSF is 0.705, and their empirical FDRs remain below the nominal levels.
OHSF and OHSF-fixed have nearly identical power.


\begin{figure}[!htbp]\centering
\includegraphics[width=116mm]{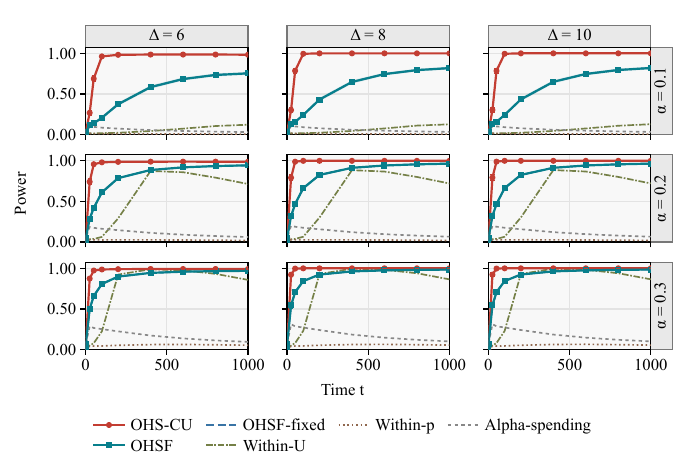}
\caption{Power at $T=1000$ with exponential spending
$\gamma_t=0.003(0.997)^{t-1}$. Rows, columns, repetitions, and intervals
follow Figure~\ref{fig:supp-cx-power}; time is shown in thousands.
}\label{fig:supp-long-power}
\end{figure}

\begin{figure}[!htbp]\centering
\includegraphics[width=116mm]{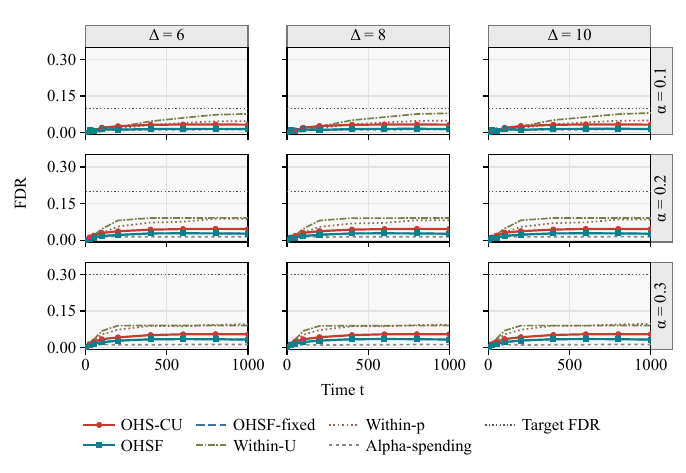}
\caption{Empirical FDR under the setting of Figure~\ref{fig:supp-long-power}.
The dotted reference is the nominal level in each row.}\label{fig:supp-long-fdr}
\end{figure}

\FloatBarrier
\paragraph{Supplementary simulation: a longer time horizon.}
We retain the original binary response, $c=0.5$, fitted-score procedure,
$K=400$, $T_0=20{,}000$, and $u=10$, but extend the horizon to $T=1000$.
After 40 separate development repetitions, we retain exponential spending
with $\gamma_t=0.003(0.997)^{t-1}$, whose infinite sum is one and whose first
1000 terms sum to approximately 0.9504, without renormalization.
We use 500 independent confirmation repetitions and the same
$(\alpha,\Delta)$ grid as above.
Figures~\ref{fig:supp-long-power}--\ref{fig:supp-long-fdr} show that the
minimum mean power of the two main methods in the nine displayed settings is
0.754, with empirical FDR below the nominal levels.
A paired reference using the original $0.05(0.95)^{t-1}$ sequence has
terminal power between 0.080 and 0.115 in these settings.
This improvement concerns expenditure across the longer horizon;
OHSF remains close to OHSF-fixed.

\FloatBarrier

\makeatletter
\setlength{\floatsep}{6pt plus 2pt minus 2pt}
\setlength{\textfloatsep}{7pt plus 2pt minus 2pt}
\setlength{\intextsep}{5pt plus 2pt minus 2pt}
\setlength{\@fptop}{0pt}
\setlength{\@fpsep}{8pt plus 2pt}
\renewcommand{\topfraction}{.95}
\renewcommand{\bottomfraction}{.85}
\renewcommand{\textfraction}{.06}
\renewcommand{\floatpagefraction}{.78}
\setcounter{topnumber}{3}
\setcounter{bottomnumber}{2}
\setcounter{totalnumber}{4}
\linespread{.95}\selectfont
\makeatother
\subsection{Additional Results on Real Data Applications}\label{app:case-new}
The PRSA classifier uses 300 boosting iterations, at most 63 leaves per tree, a learning rate of 0.06 and an $L_2$ penalty of 2.

The selected configurations use a within-group normalization variant. Let \(d\) be the number of raw positions in the within-group panel including the candidate, \(J\) the number of null positions on the candidate-null event, and \(r\) the rank cutoff. On passing the cutoff, use \(d/r\) in place of \(J/r\). The null-weighted permutation average is bounded by \((J/d)\{\min(r,J)/J\}(d/r)=\min(r,J)/r\le1\) for \(J>0\), and is zero for \(J=0\). The feedback panel counts every previously revealed same-group observation in \(d\), irrespective of its state. The rank-fraction schedule is deterministic, so the argument applies separately at each time. For the KIBA threshold sensitivity at $c=11$, use the same setup as at $c=10.5$ except for $a_t=1$.
\begin{figure}[!htbp]\centering
\includegraphics[width=116mm]{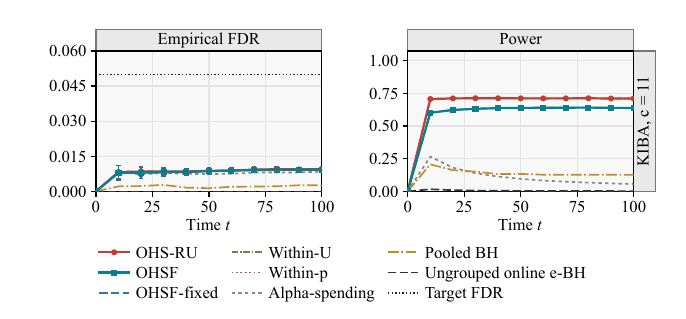}
\caption{KIBA at $c=11$: empirical FDR and power.}
\end{figure}

KIBA contains 118,254 measurements for 2,111 drugs and 229 targets \citep{ozturk2018,tang2014}. Targets define the groups, of which 174 are eligible. Use 40\% of each target's observations for external ALS training and split the remainder into 40\% development and 60\% confirmation pools. Scores are negative predicted affinities. ALS12 uses rank 12 and 25 iterations; ALS48 uses rank 48 and 35 iterations. Both use ridge penalty 8. Use the randomized revelation protocol in Section~\ref{sec:case}. The development grid crosses three thresholds, two models, two calibration sizes, three rank fractions, two horizons, three spending schedules and two FDR levels, yielding 432 configurations with 60 repetitions each. At each threshold, select the configuration maximizing the smaller terminal power of the principal procedures and run 600 confirmation repetitions (Table~\ref{tab:kiba-settings}). A power$b$ schedule normalizes $(t+4)^{-b}$.

The matched PRSA comparison fixes the model and all A22 settings except the target threshold, $c\in\{25,35,50,75\}$. Both comparisons are displayed as Power and FDR plots.

\begin{table}[H]\centering\small\caption{KIBA confirmation configurations.}\label{tab:kiba-settings}\begin{tabular}{lllllllll}\toprule
ID&$c$&Score&$m$&$u$&$q_w$&$T$&$\alpha$&Spending\\\midrule
K12&12&ALS48&10&0&0.0025&40&0.2&uniform\\
K12.5&12.5&ALS48&10&0&0.0025&40&0.2&uniform\\
K13&13&ALS48&10&0&0.0025&40&0.2&uniform\\
\bottomrule\end{tabular}\end{table}
\begin{figure}[H]\centering
\includegraphics[width=125mm]{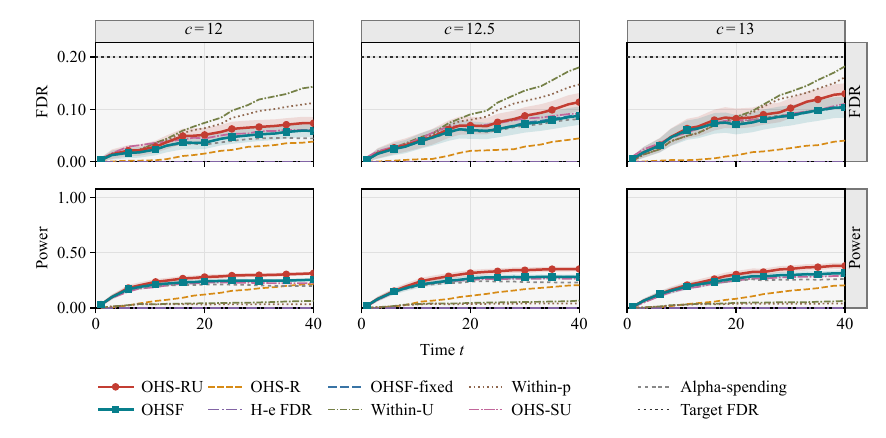}
\caption{Supplementary KIBA thresholds at \(\alpha=0.2\). Power and empirical FDR are computed from 600 paired confirmation repetitions. Shading gives pointwise 95\% Monte Carlo intervals for OHS-RU and OHSF; dotted lines mark the nominal FDR.}\label{fig:sync-kiba-Power}
\par\vspace{4mm}
\includegraphics[width=125mm]{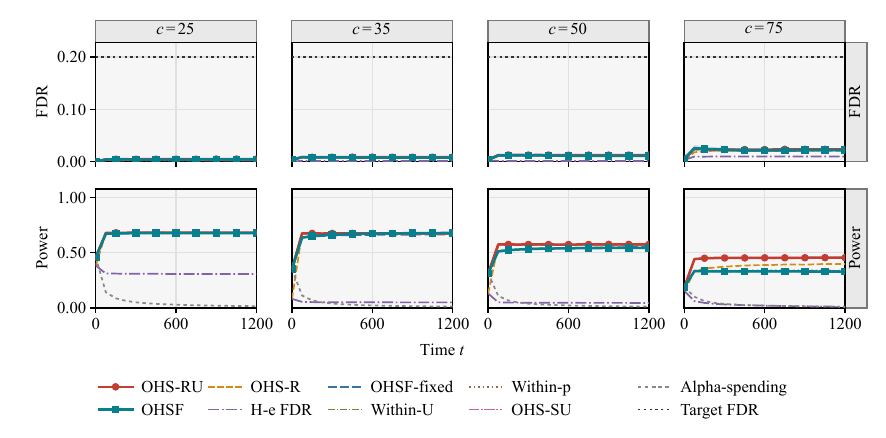}
\caption{PRSA matched-threshold comparison at \(\alpha=0.2\). Power and empirical FDR are computed from 600 paired confirmation repetitions. Shading gives pointwise 95\% Monte Carlo intervals for OHS-RU and OHSF; dotted lines mark the nominal FDR.}\label{fig:sync-matched-Power}
\end{figure}

\begin{figure}[H]\centering
\includegraphics[width=103mm]{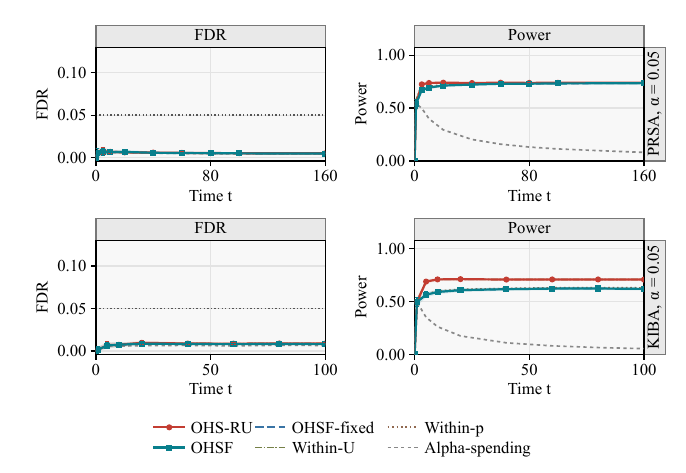}
\caption{Covariate-dependent thresholds in PRSA and KIBA at $\alpha=0.05$.
Rows correspond to datasets and columns to empirical FDR and power.
Curves average 600 paired confirmation repetitions, with pointwise 95\%
Monte Carlo intervals for OHS-RU and OHSF. Dotted references mark the nominal
FDR.}\label{fig:real-cx}
\end{figure}

\paragraph{Supplementary applications: covariate-dependent thresholds.}
We use the original external training and development/confirmation splits.
Let $\widehat\mu(X)$ be a prediction fitted on external training data: the
PRSA gradient-boosted regression prediction or the KIBA rank-48 ALS affinity
prediction. These predictions are available before observing the response.
Use
\[
 c(X)=15+2\tanh\!\left\{\frac{\widehat\mu(X)-15}{20}\right\}
 \quad\text{for PRSA},
\]
\[
 c(X)=10.5+0.5\tanh\!\left\{\widehat\mu(X)-10.5\right\}
 \quad\text{for KIBA}.
\]
PRSA retains the classifier trained at $c=15$; KIBA retains negative
predicted affinities. We use the within-group normalization described above,
with $(m,T,a_t,q_w)=(665,160,0.95,0.01)$ for PRSA and
$(155,100,0.98,0.0025)$ for KIBA, without centering.
Spending weights remain proportional to $(t+4)^{-1}$ and $(t+4)^{-1.3}$,
respectively. Eligibility requires enough observations in both splits for
the specified calibration size and horizon, leaving 12 PRSA stations and
90 KIBA targets. Following 40 development repetitions, we use 600 paired
confirmation repetitions at $\alpha=0.05$.
The varying thresholds change approximately 0.99\% and 1.53\% of the used
labels relative to 15 and 10.5, respectively.
At the terminal time, OHS-RU/OHSF power is 0.740/0.736 for PRSA and
0.706/0.618 for KIBA; the corresponding empirical FDRs are approximately
0.005/0.005 and 0.009/0.008 (Figure~\ref{fig:real-cx}).
These results demonstrate useful power under the specified varying thresholds;
they do not establish a feedback power advantage.

\begin{figure}[H]\centering
\includegraphics[width=99mm]{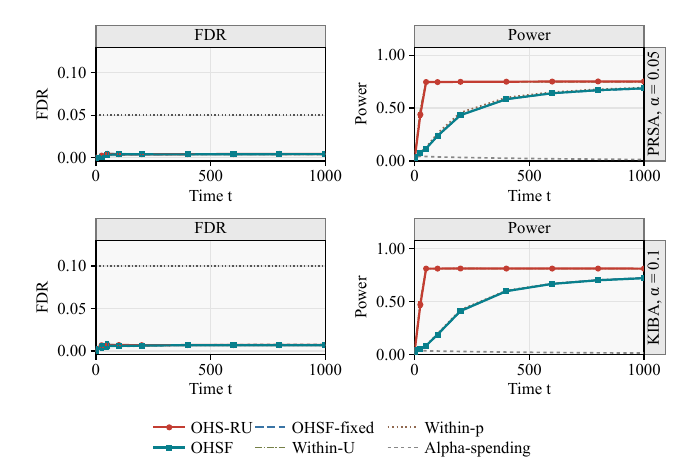}
\caption{Longer PRSA and KIBA streams with $T=1000$.
PRSA uses $\alpha=0.05$ and KIBA uses $\alpha=0.10$, as marked in the strips.
Curves and intervals follow Figure~\ref{fig:real-cx}.
Dotted references mark the dataset-specific nominal FDR.}
\label{fig:real-long}\end{figure}

\paragraph{Supplementary applications: a longer time horizon.}
We separately set $T=1000$ while retaining the original constant thresholds
$c=15$ for PRSA and $c=10.5$ for KIBA.
Use the same fitted scores and within-group normalization as above, with
$(m,a_t,q_w)=(600,0.95,0.01)$ for PRSA and $(155,0.98,0.0025)$ for KIBA.
Rechecking eligibility in both splits leaves 12 stations and 82 targets;
every repetition samples observations without replacement within each group.
We use $\gamma_t=0.003(0.997)^{t-1}$ without renormalization,
$\alpha=0.05$ for PRSA and $\alpha=0.10$ for KIBA.
After selecting settings on 40 development repetitions, 600 paired
confirmation repetitions give terminal OHS-RU/OHSF powers of 0.752/0.687
for PRSA and 0.810/0.719 for KIBA, with empirical FDRs approximately
0.004/0.004 and 0.007/0.007 (Figure~\ref{fig:real-long}).
The larger level used for KIBA is material: at $\alpha=0.05$,
OHSF power is 0.274. OHSF remains close to OHSF-fixed in the
displayed settings.

\paragraph{Supplementary PRSA stress test: unbalanced group calibration.}
Using the existing fitted regression scores, take 600 calibration records
from Dingling and 200 from Wanshouxigong, followed by $T=100$ additional
Wanshouxigong records. Records are sampled without replacement from the
confirmation split and randomly ordered within stations.
After development and candidate exploration, fix $c=75$, $\alpha=0.10$,
$q_w=0.01$, $a_t=0.95$, the raw-panel normalization, and spending weights
proportional to $(t+4)^{-1}$. OHS-RU preserves station-specific calibration pools;
the pooled baselines use the union of calibration nulls without station labels.
In 800 fresh paired repetitions, pooled BH has empirical FDR 0.114
(95\% Monte Carlo interval approximately $[0.110,0.118]$), compared with
0.046 for OHS-RU. Their powers are 0.784 and 0.471, respectively.
Ungrouped online e-BH remains below the FDR target (FDR 0.018, power 0.092).
This example shows a pooled BH FDR failure under the specified real-record
sampling protocol; OHS-RU has lower power than pooled BH.
All explored configurations and the full OHSF results are retained in the
reproducibility files.

\begin{figure}[H]\centering
\includegraphics[width=104mm]{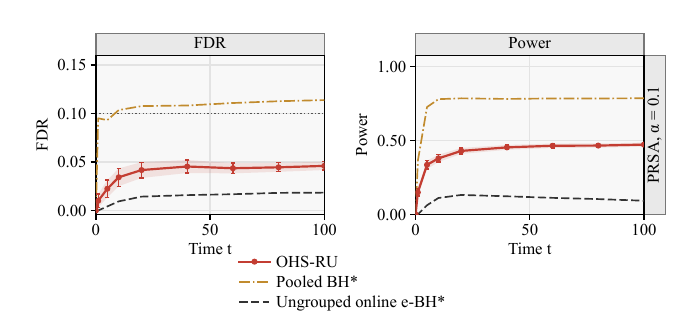}
\caption{PRSA under unbalanced station calibration: empirical FDR and power
over 800 paired repetitions. Intervals for OHS-RU are pointwise 95\% Monte Carlo
intervals. The dotted reference in the FDR panel marks $\alpha=0.10$.
Asterisks denote methods that ignore station membership.}
\label{fig:real-pooled-stress}\end{figure}

\end{document}